# Seeing Objects in a Cluttered World

Computational Objectness from Motion in Video

Douglas Poland
Lawrence Livermore National Laboratory
Livermore, California
poland1@llnl.gov

Amar Saini
Lawrence Livermore National Laboratory
Livermore, California
saini5@llnl.gov

## ABSTRACT

Perception of the visually disjoint surfaces of our cluttered world as whole objects, physically distinct from those overlapping them, is a cognitive phenomenon called objectness that forms the basis of our visual perception. Shared by all vertebrates and present at birth in humans, it enables object-centric representation and reasoning about the visual world. We present a computational approach to objectness that leverages motion cues and spatio-temporal attention using a pair of supervised spatio-temporal R(2+1)U-Nets. The first network detects motion boundaries and classifies the pixels at those boundaries in terms of their local foreground-background sense. This motion boundary sense (MBS) information is passed, along with a spatio-temporal object attention cue, to an attentional surface perception (ASP) module which infers the form of the attended object over a sequence of frames and classifies its 'pixels' as visible or obscured. The spatial form of the attention cue is flexible, but it must loosely track the attended object *which need not be visible*. We demonstrate the ability of this simple but novel approach to infer objectness from phenomenology without object models, and show that it delivers robust perception of individual attended objects in cluttered scenes, even with blur and camera shake. We show that our data diversity and augmentation minimizes bias and facilitates transfer to real video. Finally, we describe how this computational objectness capability can grow in sophistication and anchor a robust modular video object perception framework.

## 1 Introduction

Obscuration and motion are ubiquitous in the visual world. Motion of objects or sensors induces dynamics at obscuration edges called motion boundaries. Locally at these boundaries, the background object is either waxing or waning, and the foreground object remains whole. This phenomenon draws a sharp spatio-temporal visual boundary between the two objects and unambiguously identifies the local foreground/background sense of each. Thus motion boundaries provide a rich source of high precision physical object information at very low cost, meaning that no models or prior information (object type, shape, texture, etc.) are required in order to identify and interpret them. Consistent with this return on investment, motion boundary exploitation arises very early in both the phylogeny and ontegeny of biological visual perception and cognition. This work is an attempt to provide a similar foundation and return on investment for computational visual perception.

Motion phenomena obviously do not survive the projection of a visual scene into 2d, and in fact images are full of depth and object ambiguity that often goes unnoticed, but which sometimes can be jarring (Fig. 1). Parsing images into distinct objects often requires higher level cognitive processing that invokes learned models and world knowledge, effort that may be sensible as pauses or a need to shift one's gaze around during the parsing. In video however, as in the world, motion boundaries provide cues that make individual objects stand out and let us parse and attend to them effortlessly, even under extreme camouflage or in cluttered confused environments (Fig. 2). This is indicative of our inherent ability to perceive individual objects purely from motion cues. We refer to both this ability and to the spatio-temporal phenomenology that feeds it as 'objectness' in this paper. Evidence strongly suggests that objectness emerged early in vertebrate evolution [17,25], which would suggest that not only is it deeply genetically encoded, but that it may be a catalyst or even prerequisite for the acquisition and maintenance of our higher level cognitive object models [3].

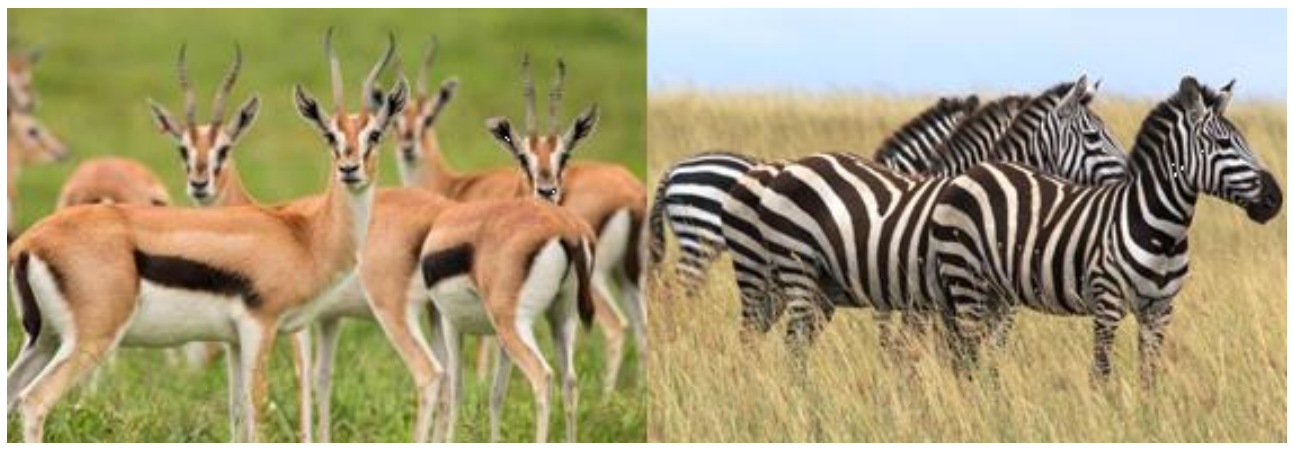

**Figure 1: Object ambiguity in images.**

Further, the ability to rapidly detect and resolve individual objects using only motion cues remains an important element of visual perception when conditions (e.g., clutter, blur, lighting) or urgency (e.g., moving through a crowd, driving in the city) limit or neutralize one's ability to cognitively attend and process every individual object based on higher level object and scene models.

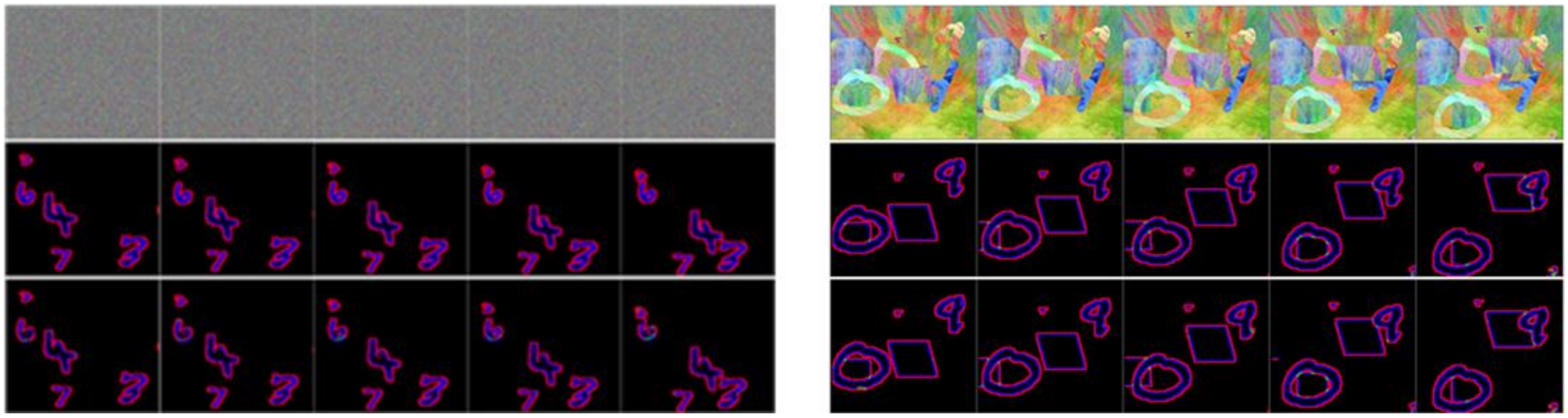

**Figure 2: Motion boundaries for generated random dot (left) and colorsplash (right) MNIST+shapes sequences. Top row: 5 frame sequence with moving digits and shapes; Middle row: MBS ground truth; Bottom row: predictions from MBSnet. Blue/red pixels denote foreground/background within 3 pixels of a motion boundary; green pixels have both senses and occur at MB intersections.**

The resolving of visual inputs and motion boundary information into individual objects, i.e. the act of visual object perception, implies that an individual object is being *attended*. This does not tell us whether this object attention is an input, an output, or an emergent phenomenon with respect to this process. For our purposes this distinction is not important, but it leads us to hypothesize that a spatio-temporal attention input function of some form must be included in ASPnet. From practical considerations, we implement object attention as a simply expressed spatio-temporal sequence (i.e., a heatmap). The key insight in ASP is that object attention should follow the spatial cohesion and temporal coherence properties of objectness and should *not* be tied to object appearance or visibility, but rather linked only to expected object form and world position (e.g., from object salience cues or prior scene knowledge). This leaves open many different options for generating this function in future work.

ASP takes as input corresponding sequences of frames, MBS predictions, and object attention, and outputs a segmentation of an attended object (Fig. 3). There is no optical flow computation, only these three inputs and a lightweight spatio-temporal U-Net.

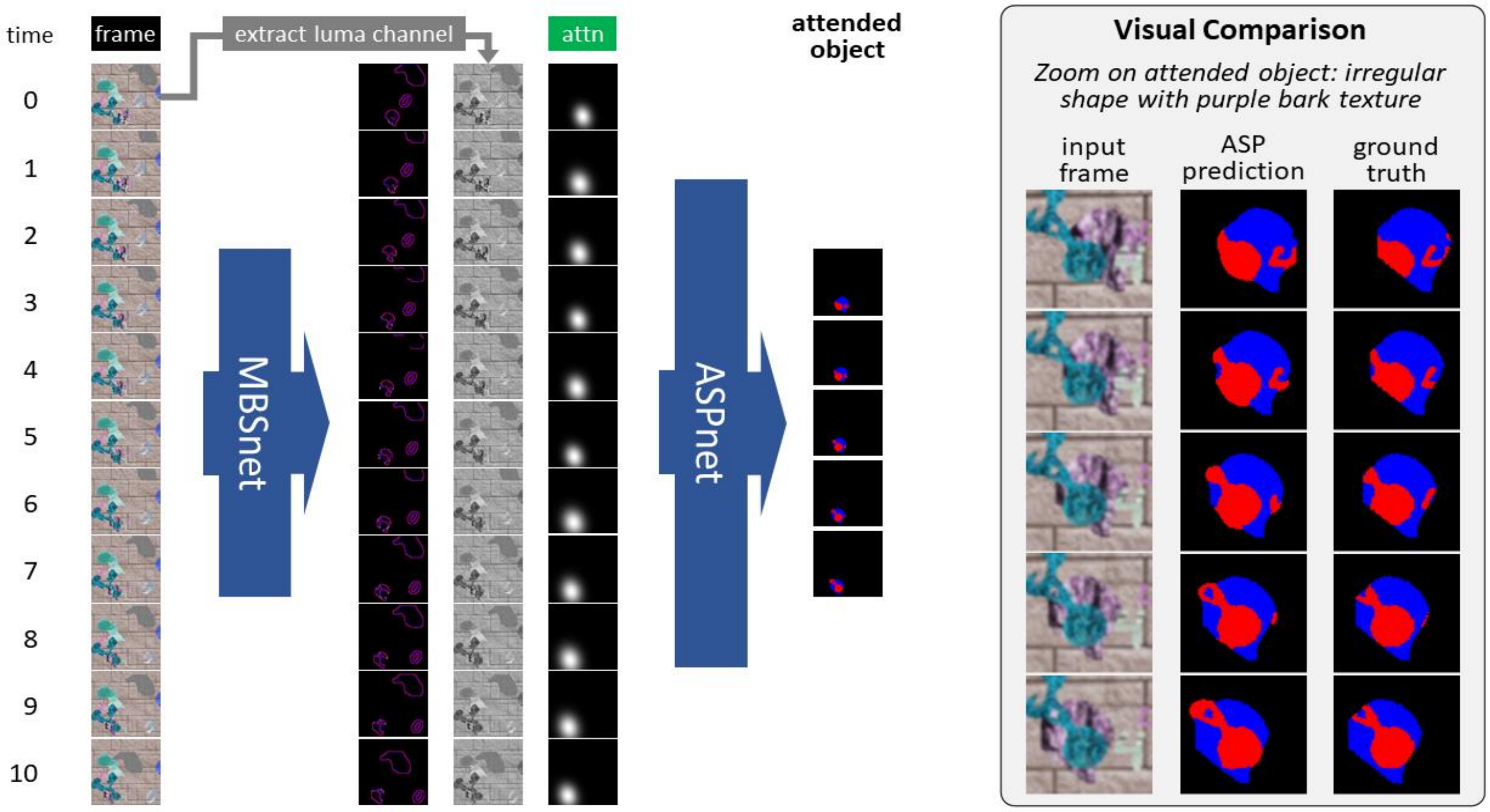

**Figure 3: Attentional Surface Perception (ASP) learns to associate a spatio-temporal attention cue with an individual object, and to infer the object's visible (blue) and obscured (red) surface over a sequence of frames. Object attention sequences may correspond to the central mass of an object as shown, to object keypoints, or to a combination. Stochastic frame blur is intentional (see Sec. 4.1).**

Our contributions include: defining the computational objectness problem; developing a local phemonenological approach to inferring MBS information directly (as opposed to first computing optical flow); developing the ASP method, and in particular demonstrating that an attention cue which itself obeys objectness is an effective way to achieve objectness in perception; and a data generation and augmentation methodology that imparts robustness to common videography induced challenges.

## 2 Objectness in Visual Cognition

Objectness and an object-centric view of the world is so deeply ingrained that it can be difficult for us to grasp that objectness is a task to be solved at all, but in the evolution of visual cognition it paved the way for everything that followed by converting incoming light to a rich, emergent, object centric representation. Biological studies across a range of species [17,25] have found these principles of visual object perception to be evident in all vertebrates: 1) Processing objects as unified entities; 2) Constant perception of invariant object properties despite changing retinal input; 3) Binding those properties in a unified representation; and 4) Attentional prioritization of object motion information. Thus objectness from motion underlies the vertebrate object perception model. These principles also emphasize early and close interaction between low-level motion processing, spatial attention, and object perception and representation. Let us further define and explore these biological capabilities and their developmental paths and functional environments in order to provide a framework for our discussion of computational objectness.

### 2.1 Objectness and the Binding Problem

From a cognitive psychology perspective, Spelke [29] discusses "one perceptual ability: the ability to organize arrays of surfaces into unitary, bounded, and persisting objects". It is noteworthy that no distinction is made between objects that are visually intact and those that are disjoint due to obscuration. The former pose a simpler foreground-background separation task (e.g. cases where moving objects are isolated against a fixed background). The latter require perceptual 'join' operations, a task referred to as either perceptual or amodal completion. We emphasize that infants have some ability to accomplish both of these tasks based primarily if not solely on motion cues, and that in both cases the result is a unitary perceived object. This genetically programmed capability is free of object models (excepting faces – see Sec. 2.2) and thus fairly naïve at birth; it is this naïve objectness that we emulate in this work.

Objects themselves are defined in terms of perception, as "something material that may be perceived by the senses" [39]. This might give one the impression that the perception of objects just happens, and in terms of *conscious* activity that is typically true. However, this ignores the very substantial and consequential challenge of binding together in a unitary object representation all of the properties derived by the various senses. Even strictly within the visual sense, binding is necessary given the many distinct visual processing pathways and centers. A clinical neuroscience analysis of the separability of various visual perception functions is presented in [18], and [10] presents an analysis of their interactions vis-à-vis binding. Treasman [33] presents a seminal description of this *binding problem* and elucidates the central role which spatial attention plays in its biological solution (though a complete understanding of that solution remains elusive). We take this binding of visual properties – form, texture, location, activity, etc. – as our practical definition of visual object perception. Objectness seems to provide the necessary foundation by producing unitary perceptual objects that correspond to physical objects. This enables the ability to purposefully attend to them and proceed to form abstract representations that properly distinguish their form, properties, and activities from those of their surroundings. In other words, we assign the ability to *abstract* a physical object to objectness, and the *construction of a unitary representation* to object perception.

### 2.2 Early Development of Object Perception

Quoting from [28], "Evolution has provided the newborn infant with the means to organise and begin to make sense of the visual world" and "they engage in very rapid learning about the visual world." The contrasts between objectness as it exists in adults, who leverage extensive learned models, versus in infants who lack object models is discussed in [29]. While there has been debate within cognitive psychology between the core-principles [29] and constructivist [13] descriptions of perceptual completion, both are consistent with the data, yielding a consensus that: 1) There are inherent core capabilities to perceive unified objects from motion, and 2) These capabilities develop more sophistication rapidly through experience. A broad review of the neuroscience of the early development of visual perception is presented in [3].

While different terms and granularity may be employed, there are parallels in the functional modularity and development of visual object perception as established in both cognitive psychology and neuroscience. These include agreement that the ontogeny of visual object perception begins: Without object models, excepting one for facelike objects; with an ability to discern objectness from motion; and on a fast learning track, ready to begin constructing and leveraging object models.

The early development of face recognition helps to illustrate these concepts and their implications. It has been shown [9] that newborns only minutes old follow moving objects, demonstrating objectness from motion as an inherent, attentive process. Their results also show that newborns respond more strongly to facelike objects, implying that newborns have a generic model for face detection. It has further been shown that at 6 months the capacity for recognizing individual faces is both modest and species independent [20]. During the first year of life, the ability to recognize individual human faces improves significantly while the ability to distinguish individuals of other primate species diminishes. Clearly this does not imply that humans cannot learn to distinguish individuals of other species, as humans learn many sophisticated identification tasks throughout their lives, it simply means that the inherent "face" model becomes tuned through experience. At the same time this speaks to the power of objectness as a module which remains intact and ready to support the learning of new categories of object models.

The scope and specifics of the interactions between visual object perception and higher level visual cognition has filled many books and continues to drive many interrelated research threads [5,6,14]. One critical theme that seems to be present in all of these is the central role of information sharing, moderated by various types of attention, between modules that perform specific inferential tasks. This seems to correlate with, if not enable, the flexibility and robustness of biological visual perception and its ability to support the learning of not one but many specialized visual cognition tasks. In fact [5] describes visual cognition as decision-based scene analysis toward the generation of representations, evoking an image of a hypothesis generation and evaluation framework. Our computational objectness framework is intended to serve as a cornerstone module in just such an evolving, flexible framework.

## 2.3 MBS and ASP as Computational Objectness

MBSnet performs one very specific low level motion processing task: direct inference of motion boundaries and the relative depth of the two adjacent surfaces. Further, we wish MBSnet to achieve this by interpreting local spatio-temporal phenomenology without *a priori* appearance models, neither from previous scenes nor even too rigidly from previous frames. This is what we mean by naïve objectness, or objectness without object models. In order to instill this in MBSnet, we strive to make the precise shape and locations of the motion boundaries unpredictable in every scene and indeed every frame through data diversity (scene unpredictability) and augmentation (frame unpredictability) as described in Section 4.1. The ability of MBSnet to robustly infer relative depth, or sense, is what makes it uniquely valuable in terms of supporting perceptual completion for partially obscured objects, and even for explaining the disappearance of fully eclipsed objects.

ASPnet combines MBSnet information with visual inputs and with a spatio-temporal attention sequence which acts as a sort of object hypothesis that drives the perception of a single object. In essence, the attention cue supervises perceptual completion. We take the attention sequence as given, though our plans are to derive it from object salience signals in future work (Section 6). To facilitate that future connection, we have imposed simple and flexible forms on these sequences (Section 4.4). As with MBSnet, data diversity and augmentation is key to ensuring that ASPnet learns perceptual completion based on spatio-temporal phenomena. Unlike MBS however, perceptual completion is not just a local perception problem and requires inferences about extended disjoint surfaces both seen and unseen.

Taken together, MBSnet and ASPnet perform a function very much like the model free, naïve objectness from motion described by [29]. Lacking biology's inherent scene executive that guides visual perception toward useful representations, we necessarily defer the connection to object representation, object salience and attention, and scene executive modules to future work (Section 6).

# 3 Related Work

The success of deep convolutional neural networks (CNNs) in solving the ImageNet Challenge (ILSVRC) [26], set in motion by [15], created huge momentum for deep CNN research. These efforts led to many successes and to revolutionary capabilities not only in image analysis and generation, but also in natural language processing, multimodal text/image problems, drug discovery and protein folding, and well posed control/gameplay challenges. However, this revolution has been far less impactful in the area of video analytics. We believe this is because video data, in general, is distinct from language and images as follows: language, whether native text or converted to text from speech, follows rules and has a limited vocabulary which makes the identification of entities and their properties straightforward. Images do not in general follow this pattern, but there is no shortage of images that are well formed enough (i.e., the subjects are centrally located and/or well set off from background and eachother) such that their recognition and segmentation is fairly straightforward. In contrast, the dynamics and diversity of video, exacerbated by imperfect videography, mean that most video content does not give up its subjects so easily.

The ILSVRC was an image classification problem, essentially a pixel region classification problem where the region was the entire image. Deep CNN solutions on ILSVRC soon spawned techniques like R-CNN [23] and YOLO [22] to find regions of interest (ROI's) in unconstrained images that lent themselves to ROI classification, extending the reach of pixel region classifiers. These have gotten traction in video object tracking and segmentation problems [35,38], in large part because CNN's are now so effective at ROI classification for objects of high interest, like people and faces. These feature-based trackers perform video object tracking by detecting and classifying ROI's one frame at a time and then linking corresponding ROI's using various association techniques [2,4]. Thus they rely on *a priori* ROI classification with learned object appearance models. As such, these are essentially object class salience cues and are complementary to objectness from motion. When available, these provide powerful cues and visual properties, but they are brittle in the presence of blur, obscuration, and novelty, all of which are ubiquitous challenges in general video.

Deep CNN's also led to new solutions for optical flow calculations [11], followed soon after by flow based inference of motion boundaries and moving object detection. Approaches producing dense optical flow are compute intensive and can produce very detailed motion boundaries [36], but they do not have depth sense and hence have much less utility for solving objectness. Flow disparity [12] or segmentation [16] is also used to detect and segment foreground objects, but it cannot distinguish individual objects. Optical flow methods have also been used to locate and interpret obscuration dynamics in order improve flow estimation in motion boundary regions in an unsupervised manner based on bidirectional census loss [19]. As an aside, we believe that this is closely related in terms of concept and function to how MBSnet solves the MBS problem, albeit MBSnet is supervised.

Finally, some recent approaches in video object tracking utilize recurrent neural networks (convGRU [31] or LSTM [27]) to instill *a posteriori* permanence in order to infer obscuration status, though they still begin with CNN based detections on single frame features. In contrast, our approach infers the physical existence (and implicit permanence) and obscuration state of visual objects via detailed spatio-temporal dynamics, without appearance models.

## 4 Approach

There are two spatio-temporal networks in our approach to computational objectness and two inputs: a frame sequence and an attention sequence (Fig. 3). We require high precision, layered object ground truth, therefore we generate our own data. Since we seek to emulate a naïve biological capability, one that achieves object and scene sophistication later in conjunction with other modules, we constrain object complexity by using only rigid (non-articulating) 2d objects. Instead we emphasize phenomenological and environmental novelty and complexity in order to produce an objectness from motion module able to support continuous learning of object and scene sophistication in real video settings.

### 4.1 Data generation and augmentation

We rely on data diversity and augmentation to eliminate object form and appearance bias to the extent possible, and also to encourage phenomenological robustness. This is a switch on the typical approach of incorporating sophisticated recognition tasks at the outset, which generally requires constraining environmental challenges like clutter, camera motion and blur. Rather we seek to achieve perception of simple but unpredictable objects in scenes that contain realistic environmental challenges. Our recipe for simulating these challenges is summarized in Figure 4.

### 4.2 R(2+1)U-Net spatio-temporal network

U-Net CNNs have proven their value on a range of practical medical image segmentation problems with various data modalities and a diverse set of target object scales and forms [24,1,8]. The environmental challenges in those domains include clutter, obscuration and blur. Inspired by these U-Net successes when faced with object form diversity and environmental challenges similar to those in visual object perception, we construct a spatio-temporal variation of a U-Net (Fig. 5). We employ the R(2+1) scheme [32], which preserves the U-Net's ability to develop multi-scale spatial strategies while facilitating synergy between the learned spatial and temporal filters. That is, it efficiently represents a world where the mapping of object motion to object texture is many-to-many.

MBS and ASP video scene diversity and augmentation

- Sample the number of moving and stationary objects
- Each object: sample shape, scale, texture, and hue
- Each moving object: sample initial velocity
- Frame to frame:
  - random blur kernel
  - frame jitter (“camera shake”)
  - gamma correction shift (whole frame)
  - hue shift (each object)
  - velocity jitter (each moving object)

ASP only

- Attention modality: central gaussian, keypoints, or both
- Frame to frame attention jitter:
  - offset relative to object/keypoint center
  - gaussian scale
- MBS signal random temporal dropout

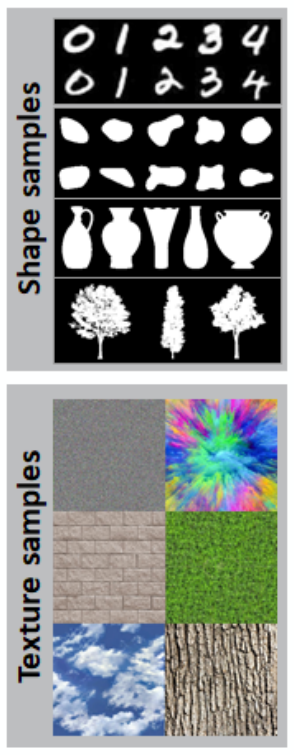

**Figure 4: Samples and pseudo-code describing our scene generation library. It produces diversity and constant motion and appearance changes which limit predictability and bias.**

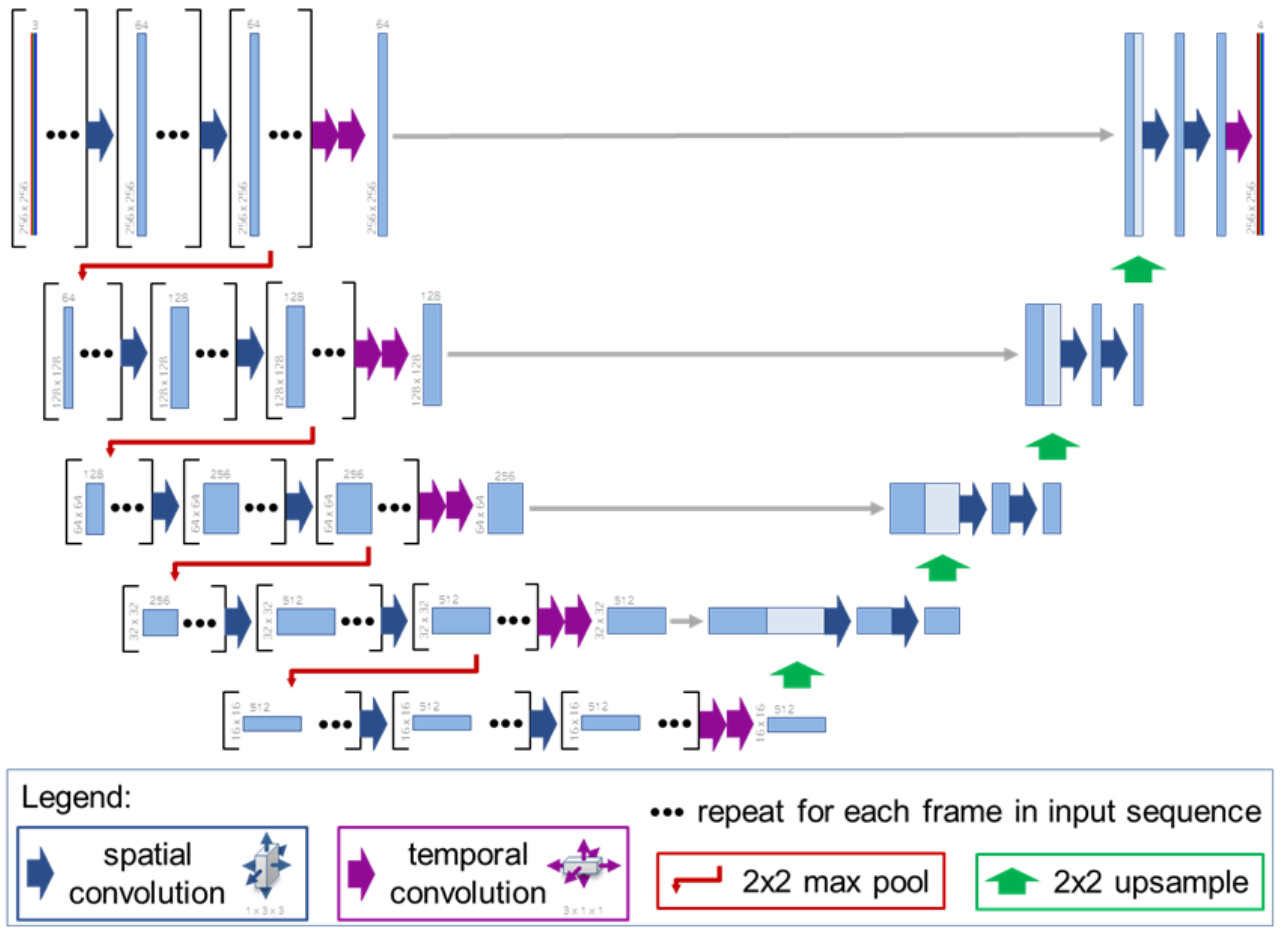

**Figure 5: The R(2+1)U-Net architecture combines [24] and [32]. MBSnet is shown; ASPnet has an additional temporal convolution at each level.**

### 4.3 MBSnet: Motion Boundary Sense Inference

Motion boundaries arise from obscuration dynamics induced by motion which yields apparent relative motion between the obscuring and obscured object. This can be induced by the motion of either object or, if there is depth disparity between the two objects, motion of the sensor. In any case, motion boundaries are realized in video as a local spatio-temporal phenomenon where the foreground object remains whole while the background object is waxing or waning. We treat this as a holistic multi-object multi-dimensional phenomenon and infer it directly with a convolutional spatio-temporal inference network (Fig. 5).

We compute training and test scenes following the recipe in Figure 4, and we compute motion boundary ground truth from the back layer forward as each new object is generated. The boundaries are expanded with morphological dilation creating a 3 pixel wide tagged band on each side of the boundary; we assign foreground pixels the blue output channel and background pixels the red channel. Where motion boundaries intersect there will be pixels with foreground with respect to one layer and background sense with respect to the other. We assign these their own class (the green output channel). The output of MBSnet is a 4 channel “softmax image’ with the same width and height as the input frames.

### 4.4 ASPnet: Attentional Surface Perception

The task of attentional surface perception is to identify and segment the surface of a single attended object into visible and obscured pixels in the presence of clutter and environmental challenges. ASPnet takes as inputs three temporal sequences: frames, MBSnet predictions, and spatial attention (Fig. 3). Solving this task – objectness from motion – may begin with local motion boundary phenomena but it also requires inferences to fill in boundaries and join multiple regions to complete the object, and then going one important step further to interpret each pixel in the inferred object as either obstruction or target. While this task is not confined to local phenomenology like MBS, it is still a multi-scale spatio-temporal inference problem. We therefore utilize the same R(2+1)U-Net architecture and we feed into it a 5 layer input sequence: one layer for each of the MBSnet blue, red and green softmax output channels, one layer for the input frame luma channel, and one for spatial attention as a heatmap. The condensation of frame inputs to a single luma channel is motivated by a desire to avoid redundancy and supported by the fact that the value of chromatic information in biological motion processing is controversial at best [7] (we also note that full color has already been exploited by MBSnet).

The attention sequence in ASP functions as an object hypothesis, driving ASPnet to select components that track with the attention sequence and hence with each other as well. This could be a weak hypothesis such as one might infer from motion boundary filtering or object class pixel segmentation, a somewhat stronger hypothesis from an object recognition cue which might include form detail, or possibly a very strong hypothesis from an established object track that includes strong keypoint predictions. While the realization of those possibilities (via independent modules managed by a scene executive, e.g.) is deferred to future work, we lay the groundwork for them here by generating two different types of spatial attention cues: a scaled central gaussian that loosely tracks the object center of mass, and constellations of 2 to 4 tight gaussians that emulate keypoint trackers. The constellations are generated by splitting the target object bounding box into halves (vertical, horizontal or diagonal) or quadrants, and then randomly selecting and tracking a point in each segment. The data generator randomly selects a mode from either: center of mass, constellation, or both. In all modes the attention scale and tracking is jittered during training, emulating noisy predictions (see examples in Fig. 9).

## 5 Results

For each experiment, we first generate a set of scenes and their associated ground truth following the data augmentation recipe. Each scene contains 15 frames of 256x256 pixels (but both MBSnet and ASPnet are fully convolutional so there are no frame size constraints). We use 15,000 scenes for training, 1000 for validation, and 200 for testing. For ASPnet training, each time a scene is used, the target object is randomly selected. For all experiments, we use the Adam optimizer, along with a Cyclic-based learning rate scheduler, OneCycleLR. For MBSnet training, we have the max and starting learning rates as 3e-3 and 3e-4 respectively; for ASPnet we use 3e-4 and 3e-5 respectively. Both the MBSnet and ASPnet are, in terms of their metrics, pixel classifiers, hence we utilize a cross-entropy loss. MBS mixed sense (green) pixels are weighted by a factor of 3 when computing the loss to compensate for the class imbalance and to emphasize their importance as markers of motion boundary intersections. ASP "visible object" (blue) and "obscured object" (red) pixels are weighted by a factor of 4 relative to the "no object" (black) pixels in the ASPnet loss for similar reasons.

We train MBSnet first, then freeze that model and use it to generate MBS predictions for ASPnet training. While our system is end-to-end, we focus on ASPnet training with a frozen MBSnet to avoid catastrophic forgetting of the MBS task.

### 5.1 MBSnet

MBSnet models consistently converge in just a few epochs and perform in the range reflected in the pixel class label confusion matrices presented in Figure 6. This reflects the conciseness of the task and the ability of the generated data (and loss metric) to convey that task effectively. While solving MBS inference on test data is an important metric, the real guage of MBSnet model utility is domain transfer to real video. We relied heavily on qualitative transfer testing to inform our data generator development. We can also assess data augmentation choices by cross testing our models on the different generated test datasets; for example, Figure 7 illustrates the dramatic importance of augmenting the data with 3x3 and 5x5 blur kernels. Model 11a was trained without blur and misses many boundary pixels on blurred data. Models 11c and 12 were trained on data with blur, while Model 11c used batch normalization during training, while Model 12 uses instance normalization [34] resulting in a boost in performance that can be seen both quantitatively (Fig. 6) and qualitatively (Fig. 7).

We present the results of testing MBSnet Model 12 on a variety of other videos in Figure 8. Despite training entirely on rigid 2d objects, this model displays impressive powers of MBS inference on blurry articulating objects, objects with aspect changes and out of plane rotation, within natural scenes and video game data that have very different pixel statistics and motion characteristics, and with substantial and diverse camera motion characteristics. This supports our claim that MBSnet is an accomplished, albeit object-naïve, interpreter of the spatio-temporal phenomenology of dynamic object occlusion.

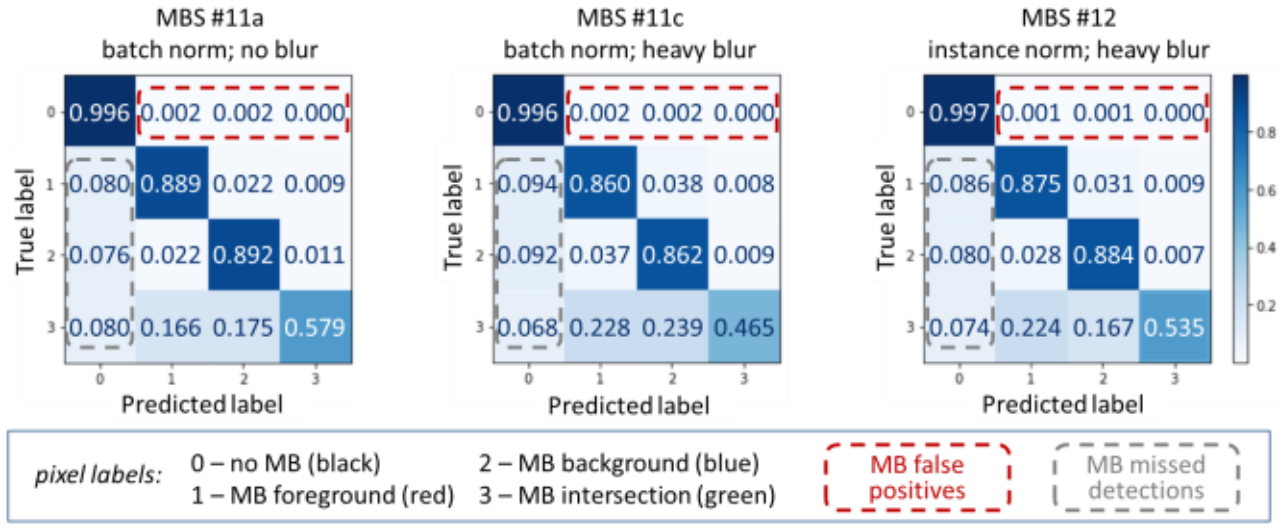

**Figure 6: Normalized confusion matrices for three MBSnet models. Due to row normalization and class imbalance, false positives are not well represented; class 3 (green) precision for these three models is 0.492, 0.488, and 0.538, respectively.**

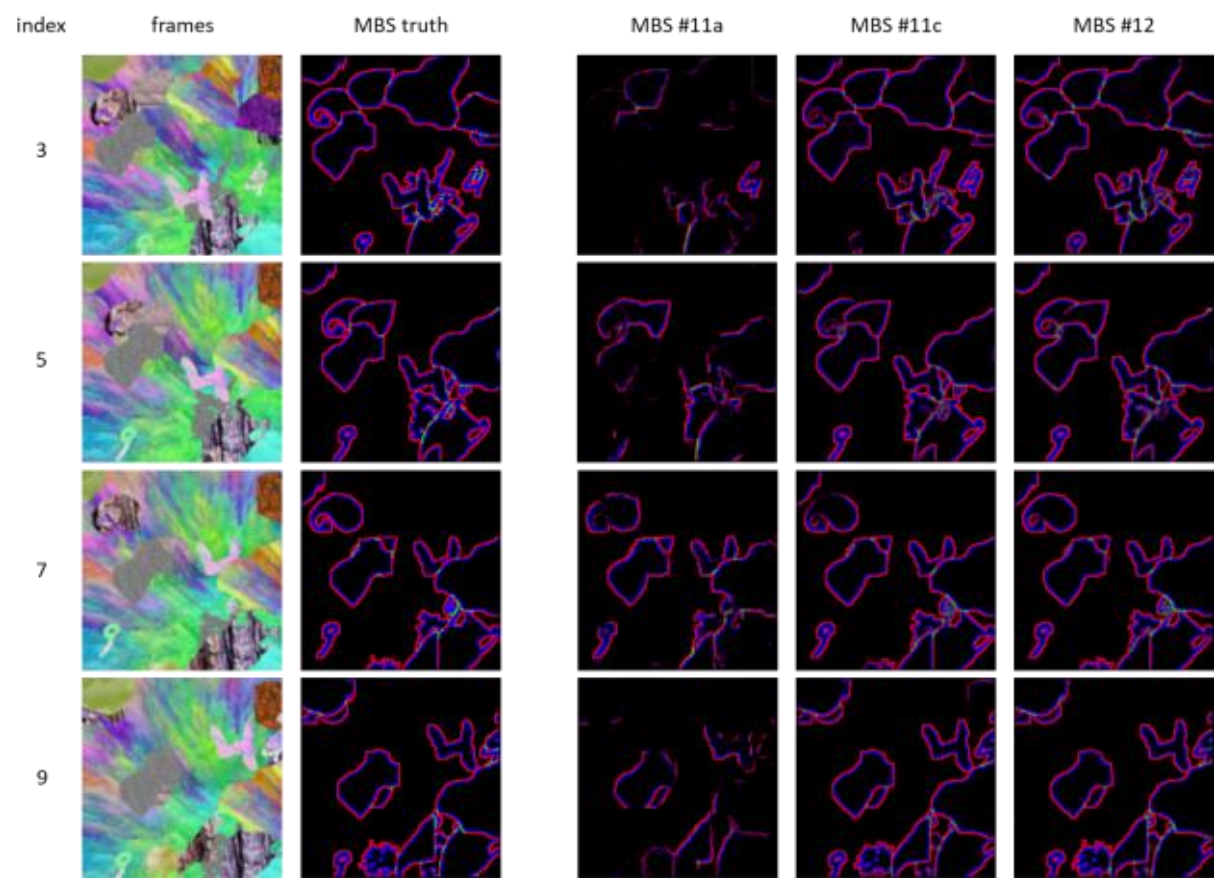

**Figure 7: Comparison of MBSnet models from Figure 6 tested on a sample sequence from the Model 12 test dataset.**

## 5.2 ASPnet

ASPnet models also converge quickly and consistently. As noted in Figure 4, ASPnet has additional augmentations corresponding to its additional input channels. We experimented with three different spatial attention modes (Fig. 9) and found the central scaled gaussian to yield slightly better results. We experimented with a variety of MBS prediction augmentations such as spatial MBS noise and dropout, but found temporal MBS dropout to be the most useful. This was implemented by randomly selecting a number of MBS prediction frames (3 to 4 on average, with a maximum of 6), and then setting them to zero. Training with temporal MBS dropout led to modest improvements in visible pixel inference and significant improvements in obscured pixel inference. Switching MBSnet's normalization layer from BatchNorm to InstanceNorm led to further improvements (Fig. 10).

In addition to spatial jitter in the attention sequence, we tried temporal attention dropout. Interestingly, the results mimicked biological 'inattentional amnesia' [37] – i.e., if the attention at time *t* was zeroed, the ASP prediction at time *t* was blank, and then it resumed with the next attention cue. MBS dropout does not have this effect – it leverages the entire sequence to make its best object inference even when 2 or 3 MBS predictions in a row are missing. Further, the spatial support of the attention does not seem critical (Fig. 9). Clearly attention in this framework warrants further investigation.

ASPnet learns the desired capability, effectively and efficiently extracting a single attended object amid clutter, and also interpreting obscuration. This demands flexible behavior that is responsive to specific object environments – even when attending to different objects in the same scene – as illustrated in Figure 11. Using the Captum GradCAM package to view layer attribution maps for the class 1 and 2 outputs, we get a view into how coherent well defined surface candidates evolve, with intriguing differences in how the two classes respond to the attention cue.

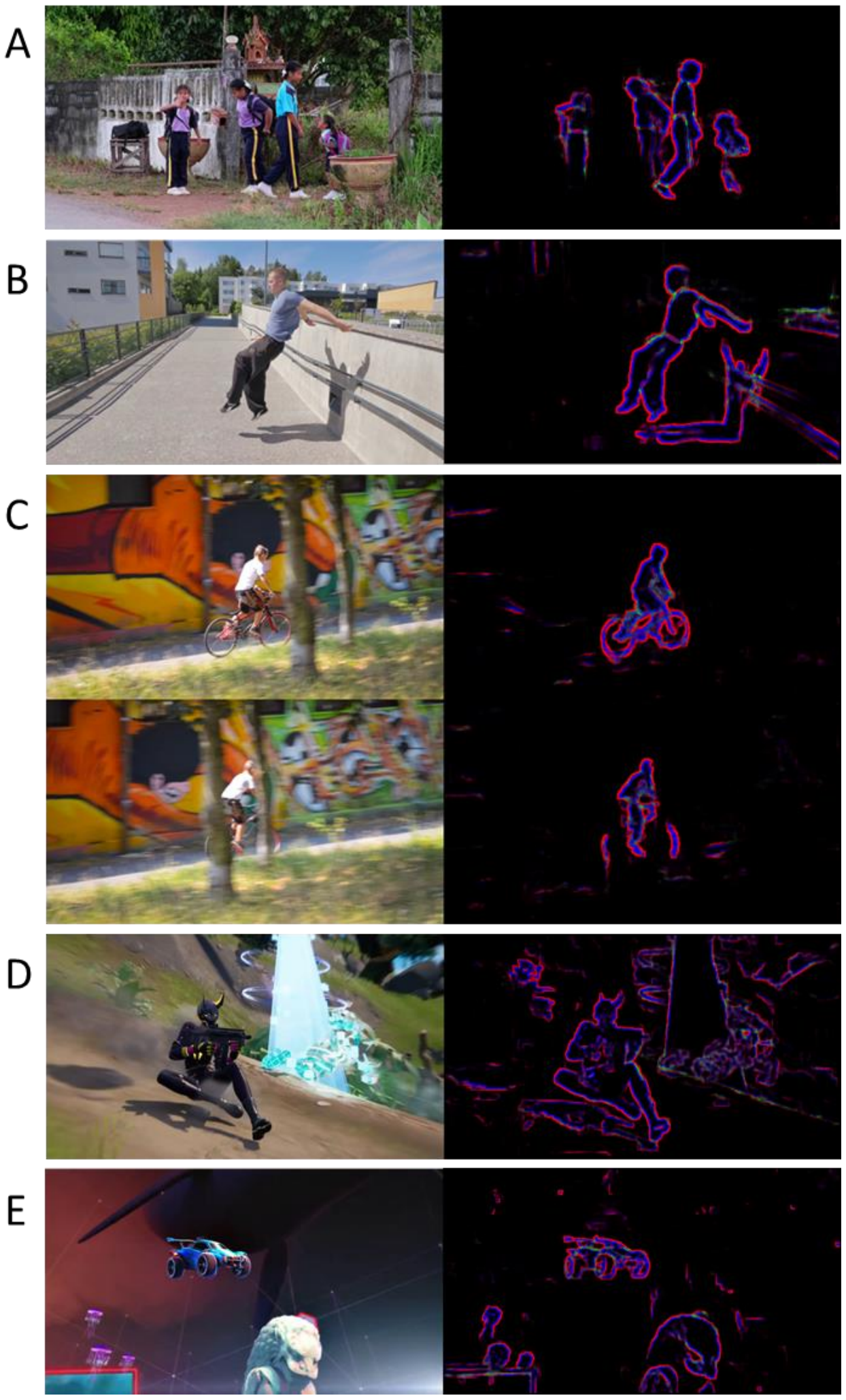

**Figure 8: MBSnet transfer test examples with diverse texture, camera motion, and blur properties. A-C from DAVIS [21]; D and E are from video games Fortnite and Rocket League.**

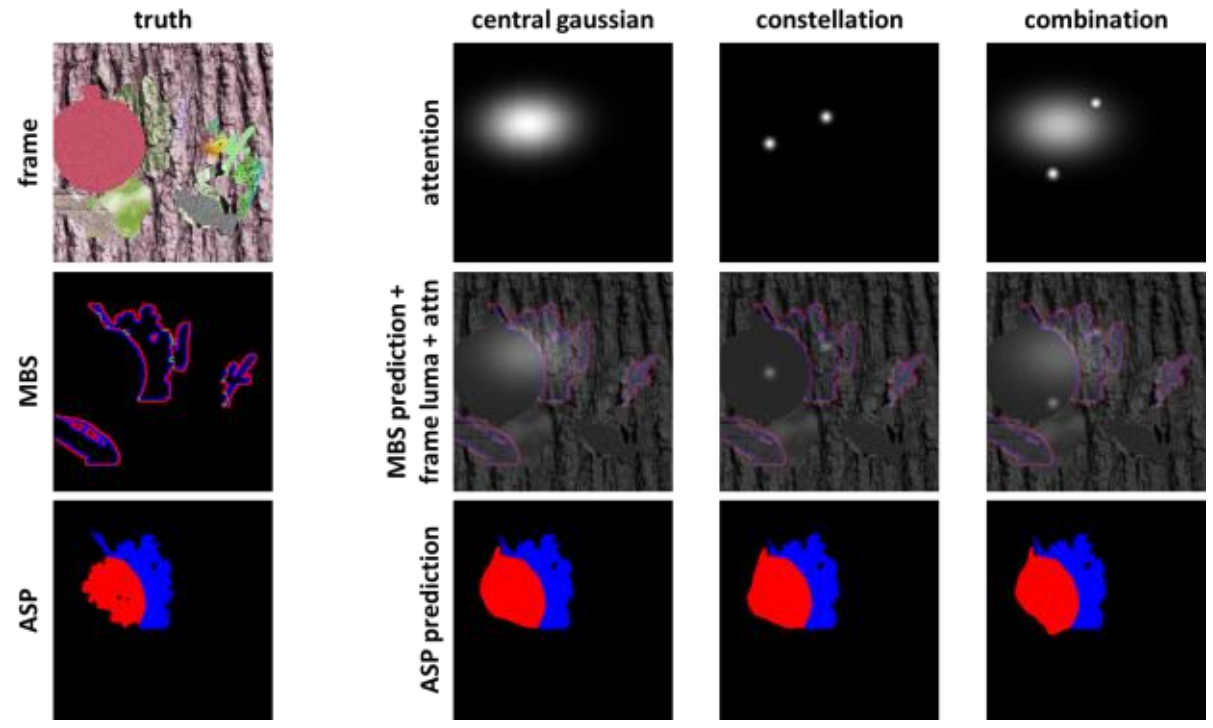

**Figure 9: ASP test sequence shapshot repeated with 3 different attention modes; ground truth is a moving tree shape with green bark texture partially occluded by a stationary shape.**

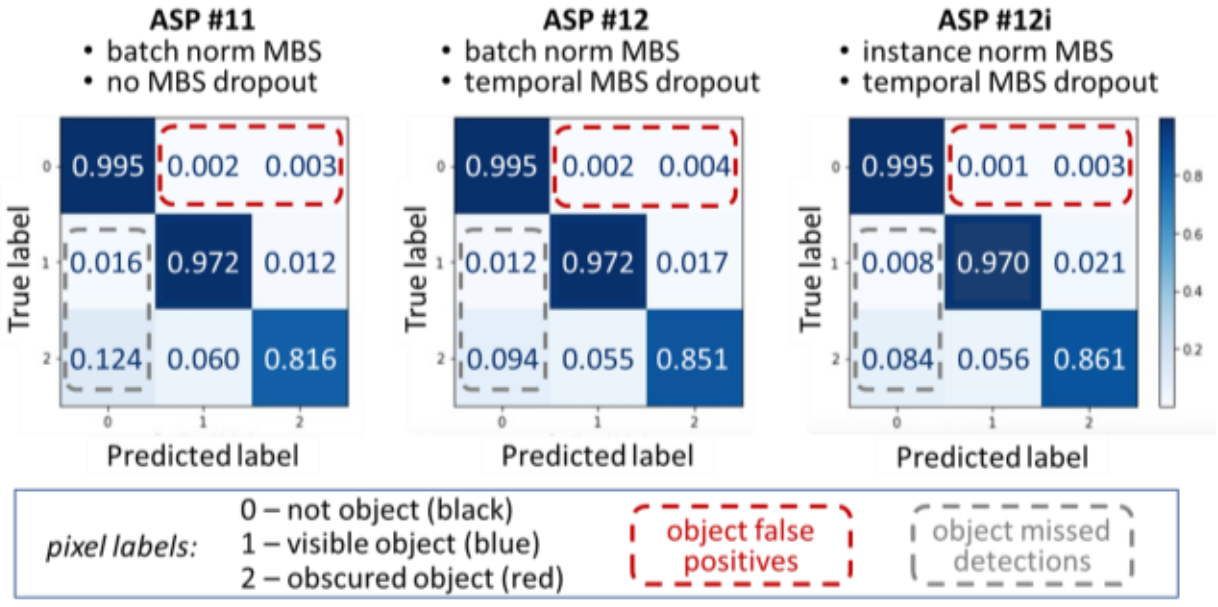

**Figure 10: ASPnet Normalized Confusion Matrices.**

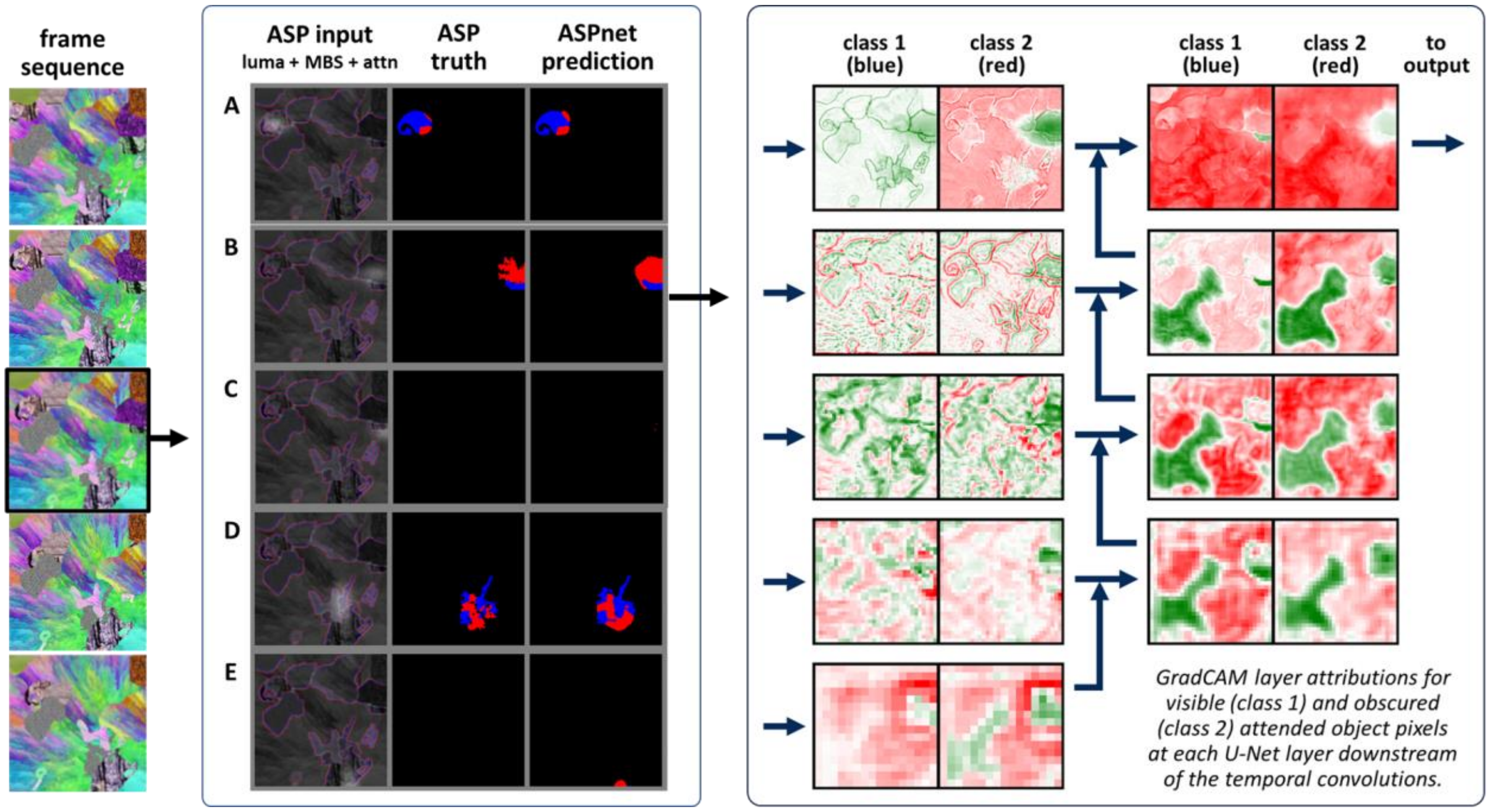

**Figure 11: Given a frame sequence (left), sequential extraction of multiple objects in the same scene with ASPnet. The luma and MBS inputs are the same in (A-E), only the attention sequence changes. The ASP input and output snapshots correspond to the middle frame of the input sequence (outlined at left). The varying obscuration dynamics for each object result in different ASP behaviors: In (A) the object and obscuration boundaries are well revealed and defined; in (B) the object is nearly entirely obscured with no information revealed about the hidden portion, so the estimate fills the obscuration footprint; in (C) the target object is offscreen in this frame, and there are no motion boundaries around the attention resulting in a "no object" inference; in (E) the target object is again offscreen, but there is an apparent obscuration resulting in a "hidden unresolved object" inference, or blob. The GradCAM maps (right) represent the influence of various layers on the final class inference for object (B); they provide a window into how ASPnet combines MBS, surface coherence, and attention. More detailed examples in Supplemental Information.**

## 6 Discussion and Future Work

We have presented an approach to computational objectness and demonstrated how the combination of motion boundaries and spatio-temporal object attention is sufficient to perceive novel objects under cluttered conditions with realistic videography challenges. This current capability is naïve in that it does not use object models and limited in that it is trained only on rigid 2d objects. Despite this limitation, MBSnet transfers remarkably well to both real video and video game sequences; conversely this preclude transfer of the current ASPnet model to full 3d objects.

We have shown that simple, flexible spatial attention sequences which loosely track an object and which obey spatial coherence and temporal cohesion, when combined with MBSnet predictions, effectively impart objectness to ASPnet. Together, MBSnet and ASPnet form a computational objectness module that is conceived and designed to grow in sophistication and flexibility as part of a larger framework, emulating the biological trajectory of visual object perception and representation [3,29,28,20]. It is ready for coupling to *separate* modules that 1) process object salience and produce object attention sequences, and 2) that leverage ASPnet object masks to produce object representations and models. Assuming those models are generative, they can produce feedback in terms of MBS boundary refinements and as attention sequences that focus on object components and/or keypoints. At the same time, there is likely to be value in exploring alternative MBSnet and ASPnet backbones as new spatio-temporal architectures emerge. Finally, we envision an executive module which will exploit the framework's object attention capability to manage and resolve various object hypotheses as it models and leverages scene context.

Achieving full 3d object and motion sophistication will require training MBSnet and ASPnet on scenes with articulating 3d objects and out of plane rotation and motion, which is feasible with video game engines. Our results with MBSnet transfer testing support our hypothesis that proper data diversity and augmentation, combined with our metrics that focus on phenomenology rather than ROI pixel pattern recognition, will yield robust transfer to real video once our networks are trained with full 3d phenomenology.

## ACKNOWLEDGMENTS

This work was performed under the auspices of the U.S. Department of Energy by Lawrence Livermore National Laboratory under Contract DE-AC52-07NA27344.

# REFERENCES

[1] Md Zahangir Alom, Mahmudul Hasan, Chris Yakopcic, Tarek M. Taha, and Vijayan K. Asari. 2018. Recurrent Residual Convolutional Neural Network based on U-Net (R2U-Net) for Medical Image Segmentation. *arXiv:1802.06955 [cs]* (May 2018). Retrieved April 5, 2022 from http://arxiv.org/abs/1802.06955

[2] Philipp Bergmann, Tim Meinhardt, and Laura Leal-Taixe. 2019. Tracking Without Bells and Whistles. In *2019 IEEE/CVF International Conference on Computer Vision (ICCV)*, IEEE, Seoul, Korea (South), 941–951. DOI:https://doi.org/10.1109/ICCV.2019.00103

[3] Oliver Braddick and Janette Atkinson. 2011. Development of human visual function. *Vision Research* 51, 13 (July 2011), 1588–1609. DOI:https://doi.org/10.1016/j.visres.2011.02.018

[4] Guillem Braso and Laura Leal-Taixe. 2020. Learning a Neural Solver for Multiple Object Tracking. In *2020 IEEE/CVF Conference on Computer Vision and Pattern Recognition (CVPR)*, IEEE, Seattle, WA, USA, 6246–6256. DOI:https://doi.org/10.1109/CVPR42600.2020.00628

[5] Patrick Cavanagh. 2011. Visual cognition. *Vision Research* 51, 13 (July 2011), 1538–1551. DOI:https://doi.org/10.1016/j.visres.2011.01.015

[6] Robert Cohen (Ed.). 1985. *The Development of spatial cognition*. L. Erlbaum Associates, Hillsdale, N.J.

[7] Simon J. Cropper and Sophie M. Wuerger. 2005. The Perception of Motion in Chromatic Stimuli. *Behavioral and Cognitive Neuroscience Reviews* 4, 3 (September 2005), 192–217. DOI:https://doi.org/10.1177/1534582305285120

[8] Kaushik Dutta. 2021. Densely Connected Recurrent Residual (Dense R2UNet) Convolutional Neural Network for Segmentation of Lung CT Images. *arXiv:2102.00663 [cs, eess]* (February 2021). Retrieved April 5, 2022 from http://arxiv.org/abs/2102.00663

[9] Carolyn C. Goren, Merrill Sarty, and Paul Y. K. Wu. 1975. Visual Following and Pattern Discrimination of Face-like Stimuli by Newborn Infants. *Pediatrics* 56, 4 (October 1975), 544–549. DOI:https://doi.org/10.1542/peds.56.4.544

[10] Stephen Grossberg. 1991. Why do parallel cortical systems exist for the perception of static form and moving form? *Perception & Psychophysics* 49, 2 (March 1991), 117–141. DOI:https://doi.org/10.3758/BF03205033

[11] Eddy Ilg, Nikolaus Mayer, Tonmoy Saikia, Margret Keuper, Alexey Dosovitskiy, and Thomas Brox. 2017. FlowNet 2.0: Evolution of Optical Flow Estimation with Deep Networks. In *2017 IEEE Conference on Computer Vision and Pattern Recognition (CVPR)*, IEEE, Honolulu, HI, 1647–1655. DOI:https://doi.org/10.1109/CVPR.2017.179

[12] Eddy Ilg, Tonmoy Saikia, Margret Keuper, and Thomas Brox. 2018. Occlusions, Motion and Depth Boundaries with a Generic Network for Disparity, Optical Flow or Scene Flow Estimation. In *Computer Vision – ECCV 2018*, Vittorio Ferrari, Martial Hebert, Cristian Sminchisescu and Yair Weiss (eds.). Springer International Publishing, Cham, 626–643. DOI:https://doi.org/10.1007/978-3-030-01258-8_38

[13] Scott P. Johnson. 2004. Development of Perceptual Completion in Infancy. *Psychological Science* 15, 11 (November 2004), 769–775. DOI:https://doi.org/10.1111/j.0956-7976.2004.00754.x

[14] Alan Kennedy (Ed.). 2000. *Reading as a perceptual process* (1st ed ed.). Elsevier, Amsterdam ; New York.

[15] Alex Krizhevsky, Ilya Sutskever, and Geoffrey E. Hinton. 2012. ImageNet Classification with Deep Convolutional Neural Networks. In *Advances in Neural Information Processing Systems 25*, Curran Associates, Inc., 1097–1105. Retrieved from https://proceedings.neurips.cc/paper/2012/file/c399862d3b9d6b76c8436e924a68c45b-Paper.pdf

[16] Runtao Liu, Zhirong Wu, Stella X. Yu, and Stephen Lin. 2021. The Emergence of Objectness: Learning Zero-Shot Segmentation from Videos. *arXiv:2111.06394 [cs]* (November 2021). Retrieved April 5, 2022 from http://arxiv.org/abs/2111.06394

[17] Elena Mascalzoni and Lucia Regolin. 2011. Animal visual perception. *WIREs Cogn Sci* 2, 1 (January 2011), 106–116. DOI:https://doi.org/10.1002/wcs.97

[18] Heath E. Matheson and Patricia A. McMullen. 2010. Neuropsychological dissociations between motion and form perception suggest functional organization in extrastriate cortical regions in the human brain. *Brain and Cognition* 74, 2 (November 2010), 160–168. DOI:https://doi.org/10.1016/j.bandc.2010.07.009

[19] Simon Meister, Junhwa Hur, and Stefan Roth. 2017. UnFlow: Unsupervised Learning of Optical Flow with a Bidirectional Census Loss. *arXiv:1711.07837 [cs]* (November 2017). Retrieved April 5, 2022 from http://arxiv.org/abs/1711.07837

[20] Olivier Pascalis, Michelle de Haan, and Charles A. Nelson. 2002. Is Face Processing Species-Specific During the First Year of Life? *Science* 296, 5571 (May 2002), 1321–1323. DOI:https://doi.org/10.1126/science.1070223

[21] F. Perazzi, J. Pont-Tuset, B. McWilliams, L. Van Gool, M. Gross, and A. Sorkine-Hornung. 2016. A Benchmark Dataset and Evaluation Methodology for Video Object Segmentation. In *2016 IEEE Conference on Computer Vision and Pattern Recognition (CVPR)*, IEEE, Las Vegas, NV, 724–732. DOI:https://doi.org/10.1109/CVPR.2016.85

[22] Joseph Redmon, Santosh Divvala, Ross Girshick, and Ali Farhadi. 2016. You Only Look Once: Unified, Real-Time Object Detection. In *2016 IEEE Conference on Computer Vision and Pattern Recognition (CVPR)*, IEEE, Las Vegas, NV, USA, 779–788. DOI:https://doi.org/10.1109/CVPR.2016.91

[23] Shaoqing Ren, Kaiming He, Ross Girshick, and Jian Sun. 2017. Faster R-CNN: Towards Real-Time Object Detection with Region Proposal Networks. *IEEE Trans. Pattern Anal. Mach. Intell.* 39, 6 (June 2017), 1137–1149. DOI:https://doi.org/10.1109/TPAMI.2016.2577031

[24] Olaf Ronneberger, Philipp Fischer, and Thomas Brox. 2015. U-Net: Convolutional Networks for Biomedical Image Segmentation. In *Medical Image Computing and Computer-Assisted Intervention – MICCAI 2015*, Nassir Navab, Joachim Hornegger, William M. Wells and Alejandro F. Frangi (eds.). Springer International Publishing, Cham, 234–241. DOI:https://doi.org/10.1007/978-3-319-24574-4_28

[25] Orsola Rosa Salva, Valeria Anna Sovrano, and Giorgio Vallortigara. 2014. What can fish brains tell us about visual perception? *Front. Neural Circuits* 8, (September 2014). DOI:https://doi.org/10.3389/fncir.2014.00119

[26] Olga Russakovsky, Jia Deng, Hao Su, Jonathan Krause, Sanjeev Satheesh, Sean Ma, Zhiheng Huang, Andrej Karpathy, Aditya Khosla, Michael Bernstein, Alexander C. Berg, and Li Fei-Fei. 2015. ImageNet Large Scale Visual Recognition Challenge. *Int J Comput Vis* 115, 3 (December 2015), 211–252. DOI:https://doi.org/10.1007/s11263-015-0816-y

[27] Aviv Shamsian, Ofri Kleinfeld, Amir Globerson, and Gal Chechik. 2020. Learning Object Permanence from Video. In *Computer Vision – ECCV 2020*, Andrea Vedaldi, Horst Bischof, Thomas Brox and Jan-Michael Frahm (eds.). Springer International Publishing, Cham, 35–50. DOI:https://doi.org/10.1007/978-3-030-58517-4_3

[28] A. Slater and Rachel Kirby. 1998. Innate and learned perceptual abilities in the newborn infant. *Experimental Brain Research* 123, 1–2 (October 1998), 90–94. DOI:https://doi.org/10.1007/s002210050548

[29] Elizabeth S. Spelke. 1990. Principles of Object Perception. *Cognitive Science* 14, 1 (January 1990), 29–56. DOI:https://doi.org/10.1207/s15516709cog1401_3

[30] Pavel Tokmakov, Karteek Alahari, and Cordelia Schmid. 2017. Learning Motion Patterns in Videos. In *2017 IEEE Conference on Computer Vision and Pattern Recognition (CVPR)*, IEEE, Honolulu, HI, 531–539. DOI:https://doi.org/10.1109/CVPR.2017.64

[31] Pavel Tokmakov, Jie Li, Wolfram Burgard, and Adrien Gaidon. 2021. Learning to Track with Object Permanence. *arXiv:2103.14258 [cs]* (September 2021). Retrieved April 5, 2022 from http://arxiv.org/abs/2103.14258

[32] Du Tran, Heng Wang, Lorenzo Torresani, Jamie Ray, Yann LeCun, and Manohar Paluri. 2018. A Closer Look at Spatiotemporal Convolutions for Action Recognition. In *2018 IEEE/CVF Conference on Computer Vision and Pattern Recognition*, IEEE, Salt Lake City, UT, 6450–6459. DOI:https://doi.org/10.1109/CVPR.2018.00675

[33] Anne Treisman. 1998. Feature binding, attention and object perception. *Phil. Trans. R. Soc. Lond. B* 353, 1373 (August 1998), 1295–1306. DOI:https://doi.org/10.1098/rstb.1998.0284

[34] Dmitry Ulyanov, Andrea Vedaldi, and Victor Lempitsky. 2017. Instance Normalization: The Missing Ingredient for Fast Stylization. *arXiv:1607.08022 [cs]* (November 2017). Retrieved April 5, 2022 from http://arxiv.org/abs/1607.08022

[35] Paul Voigtlaender, Michael Krause, Aljosa Osep, Jonathon Luiten, Berin Balachandar Gnana Sekar, Andreas Geiger, and Bastian Leibe. 2019. MOTS: Multi-Object Tracking and Segmentation. *arXiv:1902.03604 [cs]* (April 2019). Retrieved April 5, 2022 from http://arxiv.org/abs/1902.03604

[36] Philippe Weinzaepfel, Jerome Revaud, Zaid Harchaoui, and Cordelia Schmid. 2015. Learning to detect Motion Boundaries. In *2015 IEEE Conference on Computer Vision and Pattern Recognition (CVPR)*, IEEE, Boston, MA, USA, 2578–2586. DOI:https://doi.org/10.1109/CVPR.2015.7298873

[37] Jeremy M. Wolfe. 1997. *Inattentional amnesia: (536982012-171)*. American Psychological Association. DOI:https://doi.org/10.1037/e536982012-171

[38] Rui Yao, Guosheng Lin, Shixiong Xia, Jiaqi Zhao, and Yong Zhou. 2020. Video Object Segmentation and Tracking: A Survey. *ACM Trans. Intell. Syst. Technol.* 11, 4 (August 2020), 1–47. DOI:https://doi.org/10.1145/3391743

[39] Definition of OBJECT. Retrieved April 5, 2022 from https://www.merriam-webster.com/dictionary/object

## SUPPLEMENTAL INFORMATION

### List of supplemental video files (available on request):

1. **random_dots.gif –** Four MBS test sequences where all textures are from random dot palette
   (columns: input frame / MBS ground truth / MBSnet prediction / prediction overlaid on input)
2. **splash.mp4 –** MBS test sequence where all textures are from the color_splash palette
   (columns: input frame / MBS ground truth / MBSnet prediction / prediction overlaid on input)
3. **dog_agility.mp4** – MBS transfer test movie with DAVIS dog_agility sequence
4. **parkour.mp4** – MBS transfer test movie with DAVIS parkour sequence
5. **bmx_trees.mp4** – MBS transfer test movie with DAVIS bmx_trees sequence
6. **ASP_test_grid.gif** – ASP test sequence with a different object attended in each row

   While the input sequence has 15 frames, there are only MBSnet predictions for the center 11; those 11 frames and those predictions are included in this gif file. Since ASPnet has three temporal convolutions, there are only ASPnet predictions for the center 5 of these 11 frames. For each row, the column contents are:

   1. input frame
   2. attention
   3. input frame luma
   4. MBSnet prediction
   5. MBS ground truth
   6. overlaid ASPnet inputs (luma + MBSnet prediction + attention)
   7. ASPnet ground truth
   8. ASPnet prediction
   9. ASPnet prediction overlaid on input frame

## A. Data generator building blocks

We begin with the MNIST character set, since it will be helpful for later object representation and recognition work to have a set of object classes with many unique instances of each. Obtained from http://pytorch.org/vision/master/generated/torchvision.datasets.MNIST.html.

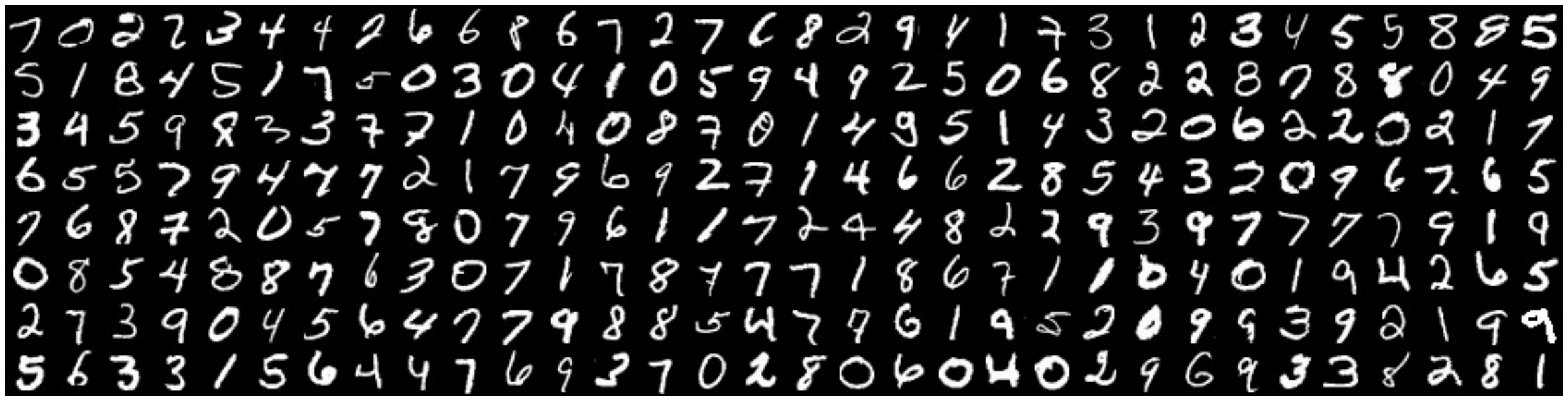

Through the course of transfer testing with MBSnet, we found that we needed more shape variety, particularly shapes with more volume, in our training set. We collected sets of shape silhouettes to add to our object candidates.

Blobs: 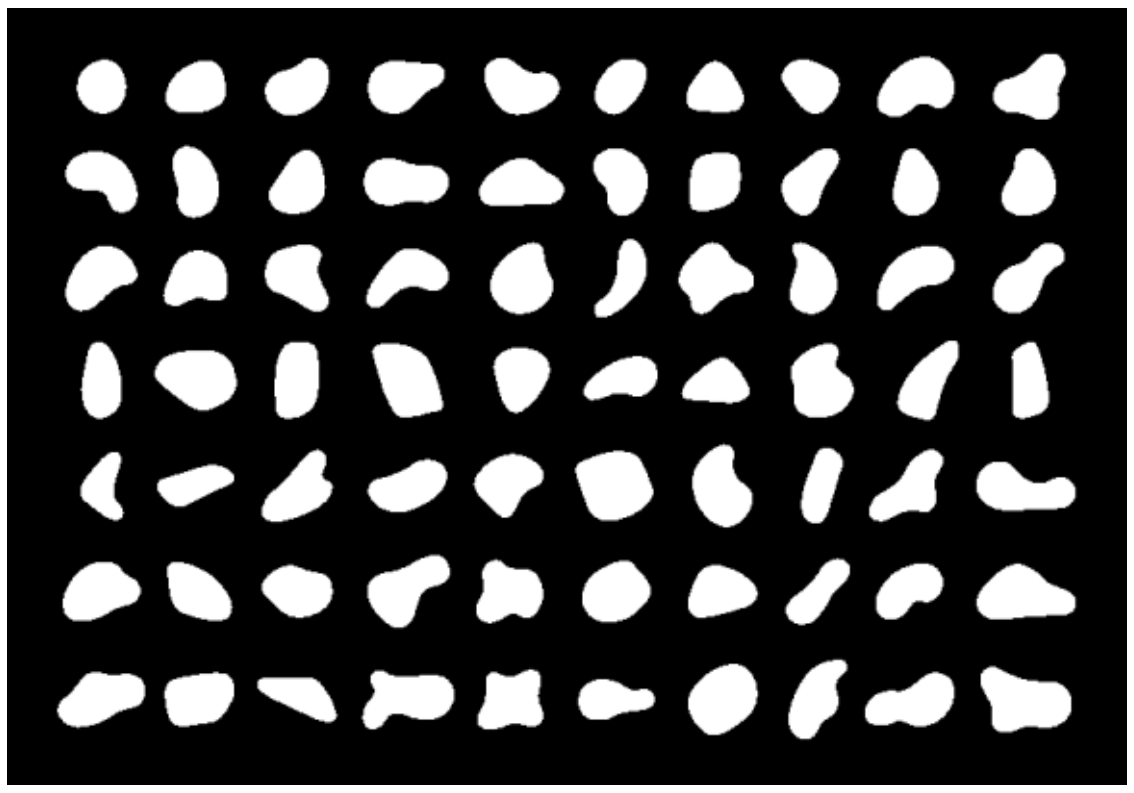

Vases: 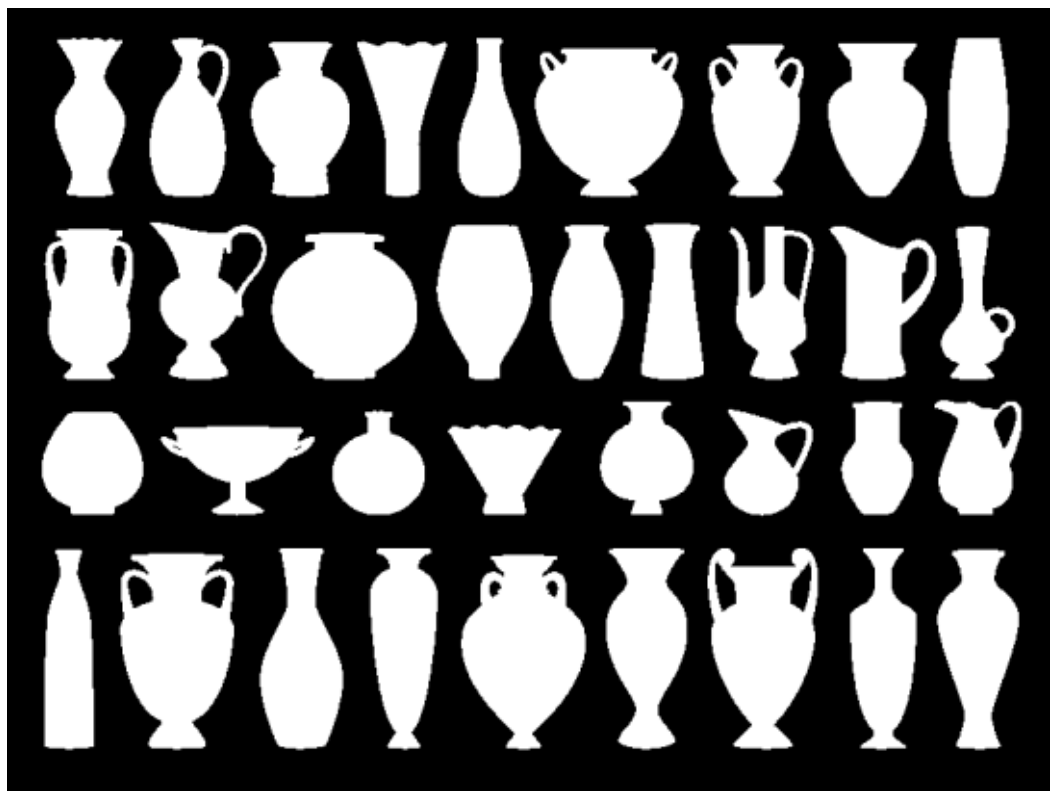

Trees: 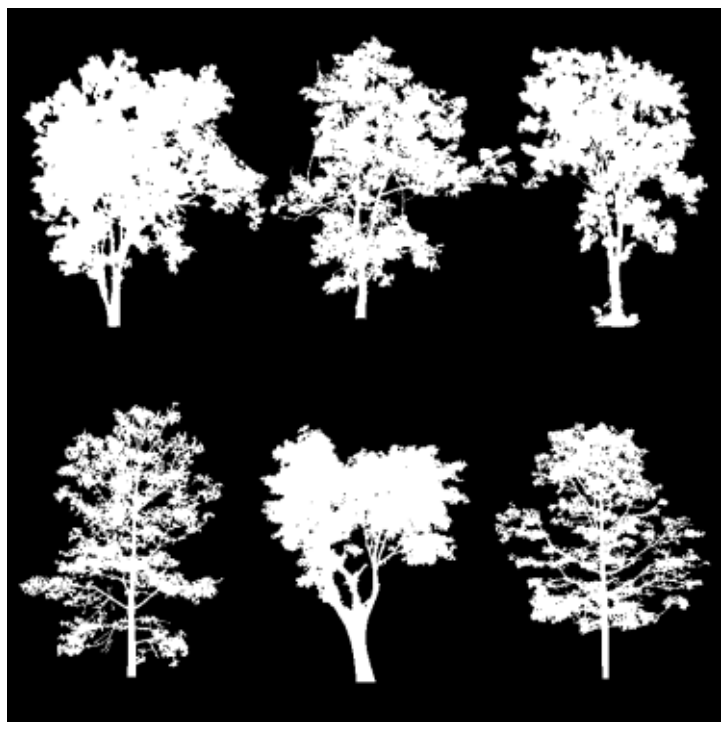

Clothes: 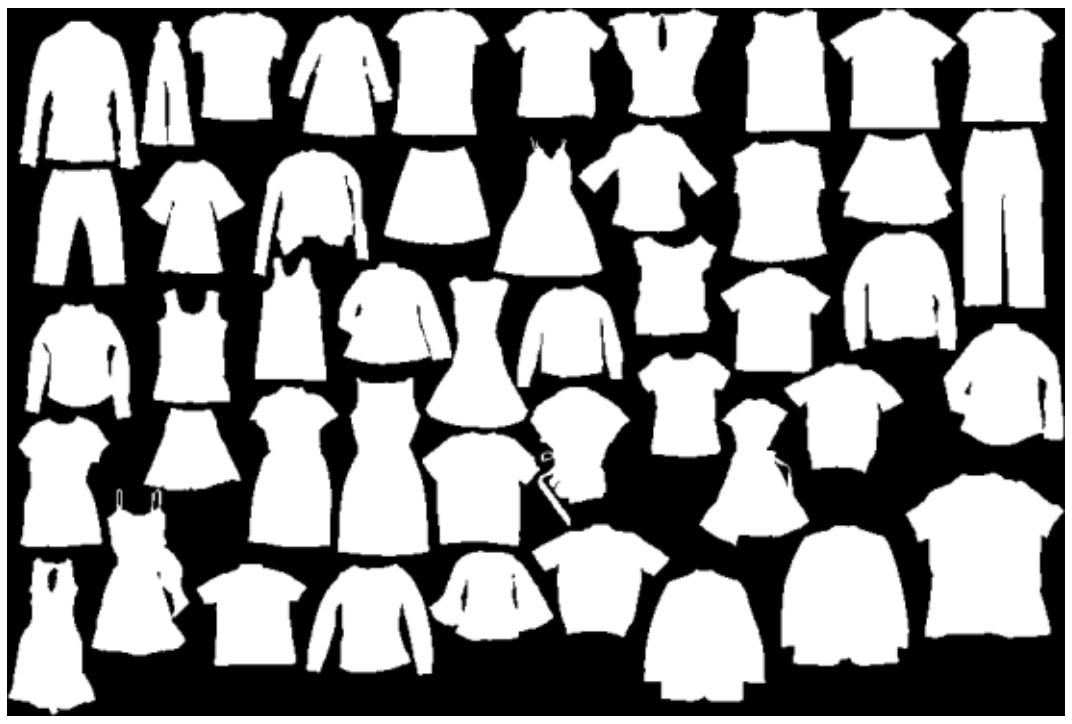

Cars: 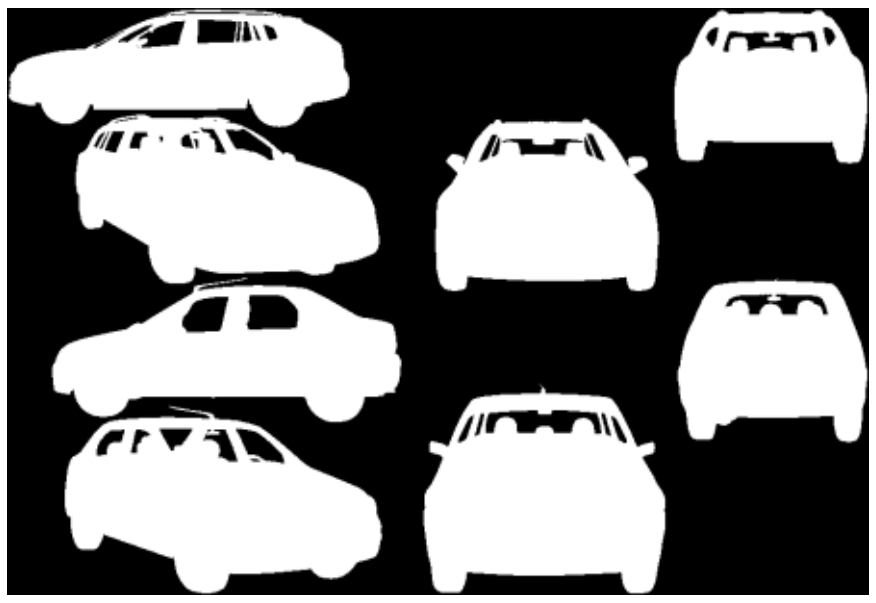

Droplets: 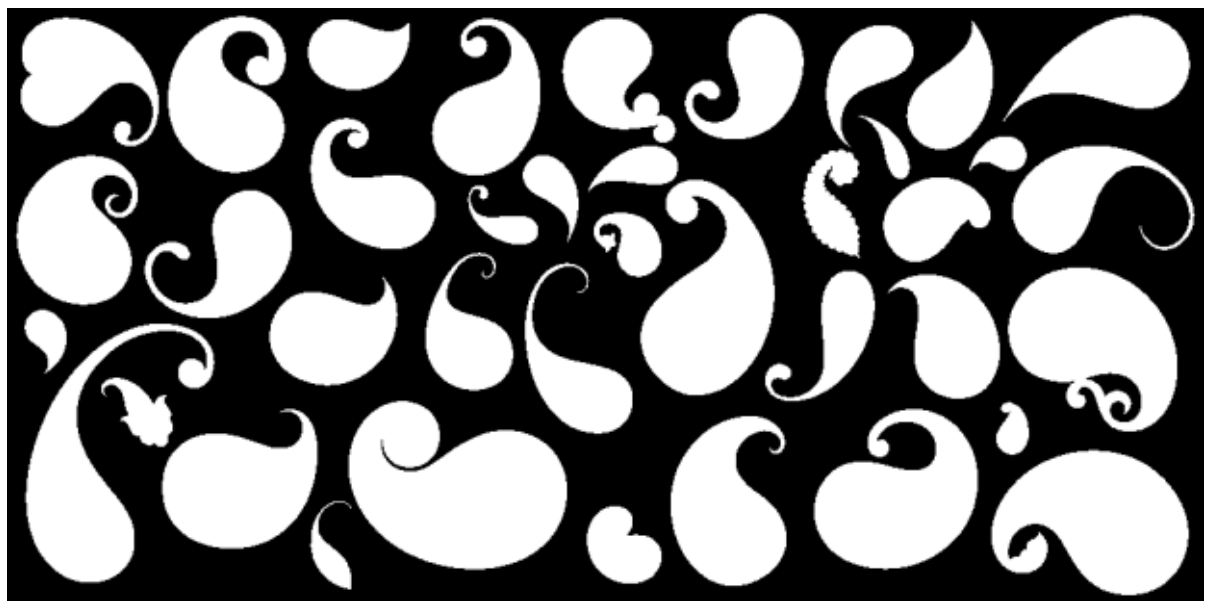

Fruit: 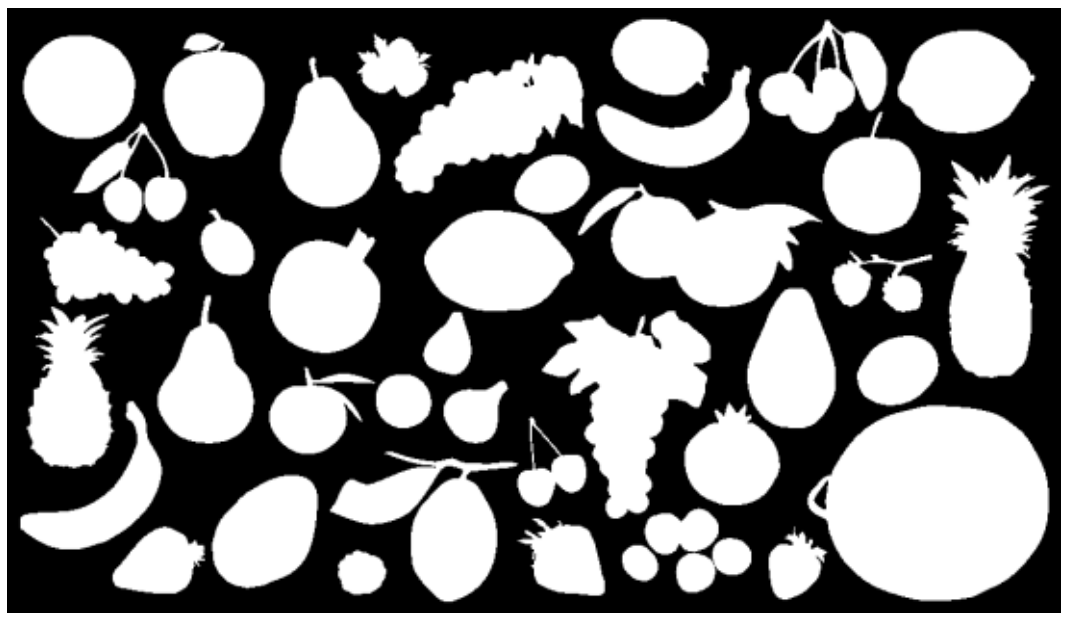

Hats: 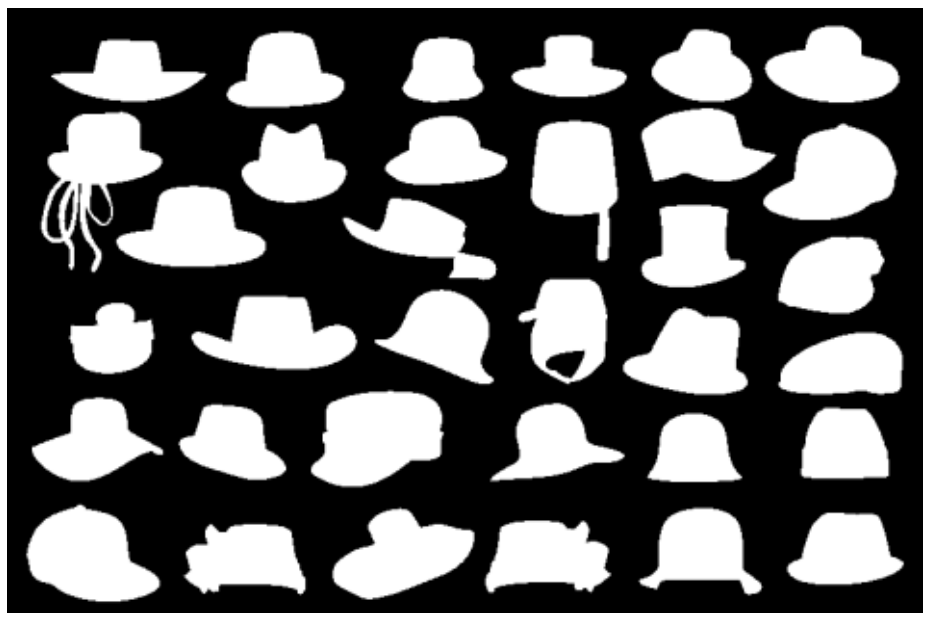

After sampling a shape and its scale and rotation, a fill texture was produced by randomly selecting one of these six palettes, sampling a location within the palette, and then randomly varying the hue:

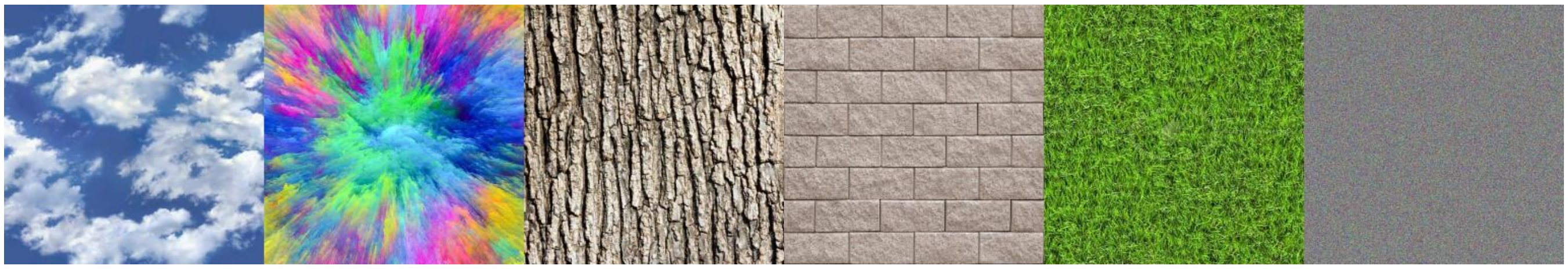

Sample snapshots of the resulting random scenes:

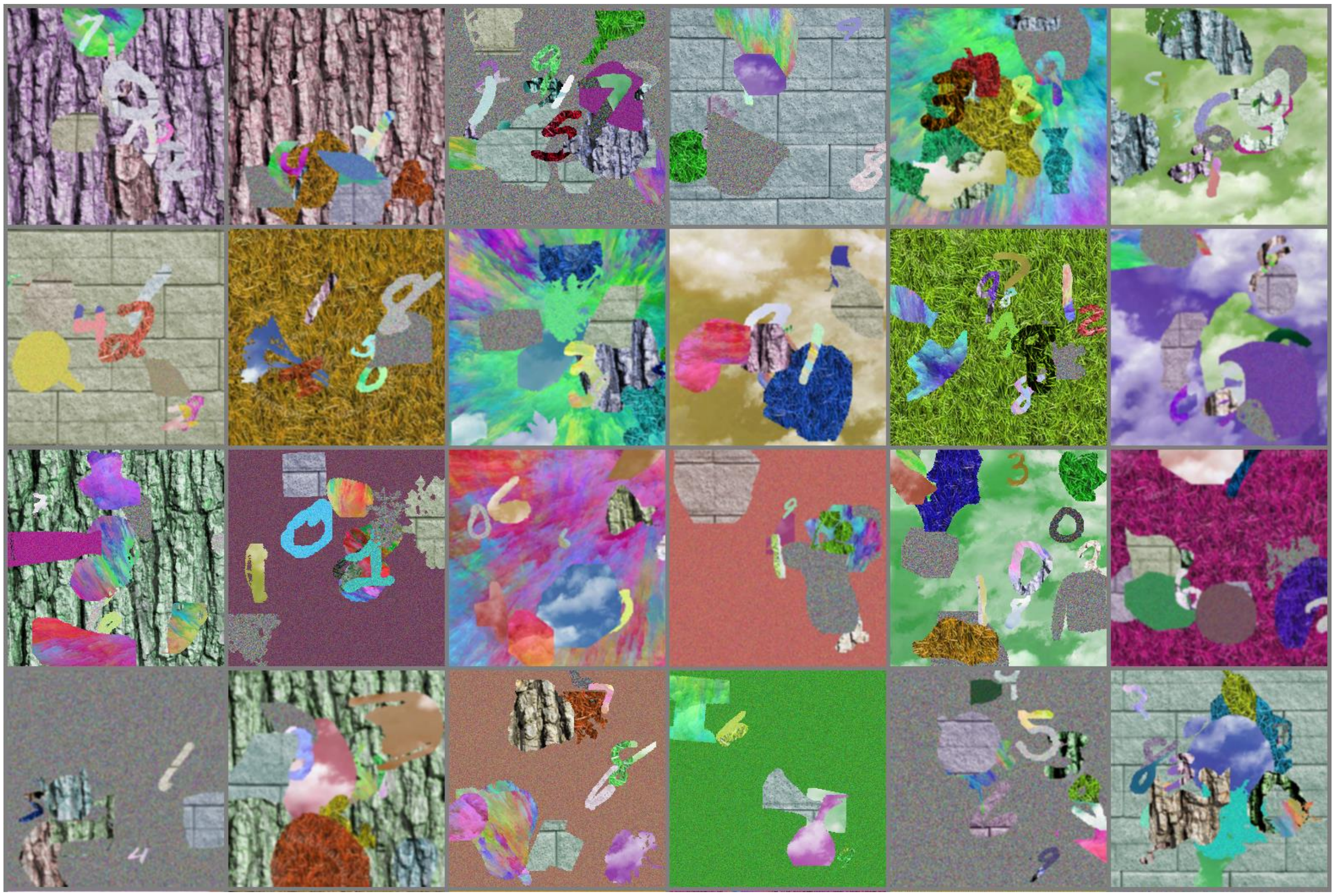

We also generated scenes where objects could have “stickers”, which were other objects that were overlayed on their texture, so these overlays would move with the host object, just like stickers or signs on a vehicle. We did qualitatively see improvement in MBSnet transfer test results when trained with stickers:

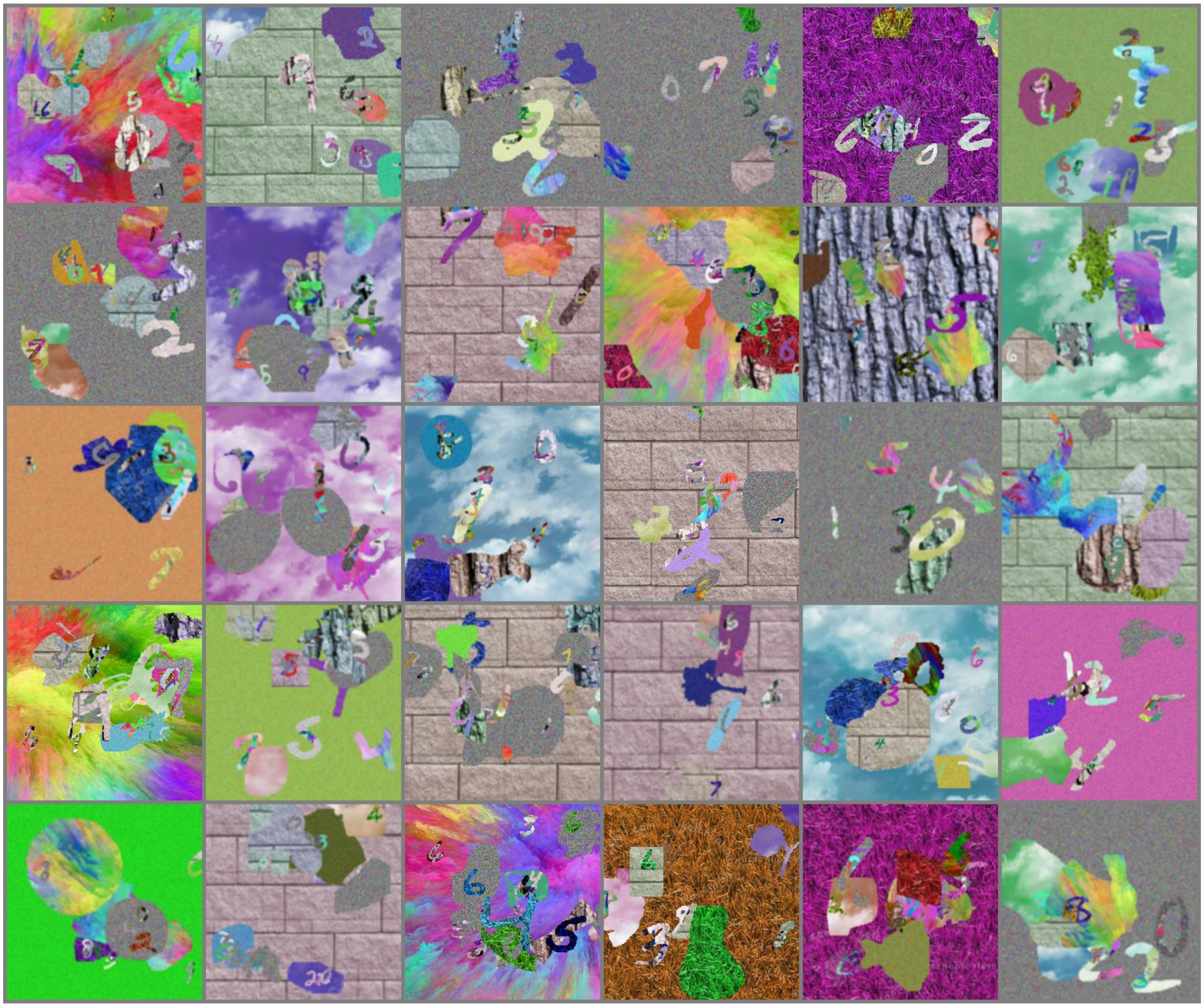

## B. MBSnet validation and test examples

Here we present a visual comparison of MBSnet performance on generated test data when trained and tested without blur (left) and with blur kernels randomly applied during generation of training and test data (right). We note that blur appears to have the greatest impact on the random dot texture objects and backgrounds (row 4 below, e.g.), which makes sense due to the complete lack of medium or large scale features present in those. While blur results in a slight drop in validation and test results, it produces transformational results in transfer testing, where motion blur is ubiquitous. When viewing these as snapshots, it is impossible to infer what is happening in the scene just by viewing the frame; we rely on the 'MBS truth' to shed light on which objects are moving and which are stationary.

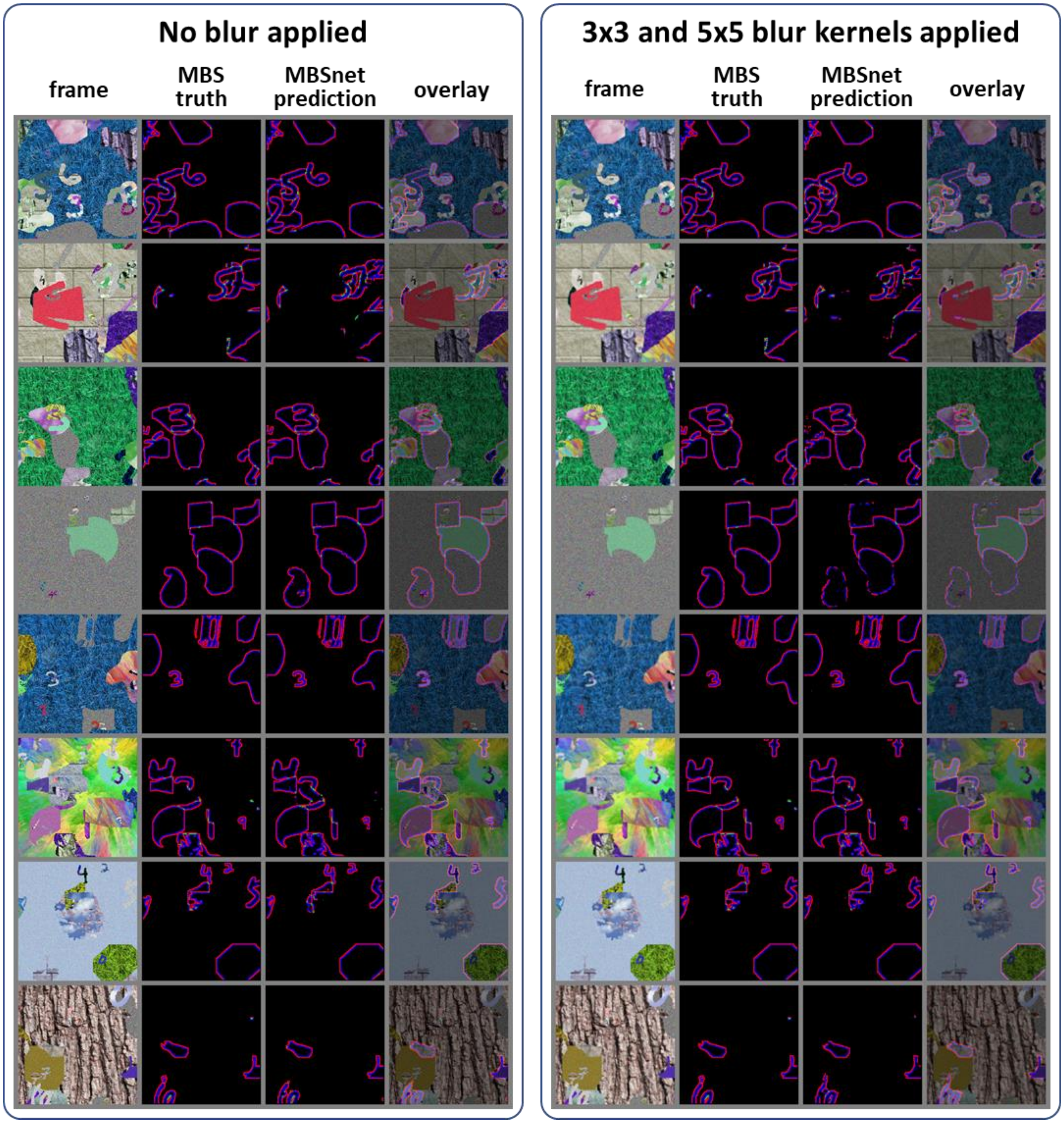

## C. MBSnet transfer examples

Fortnite:

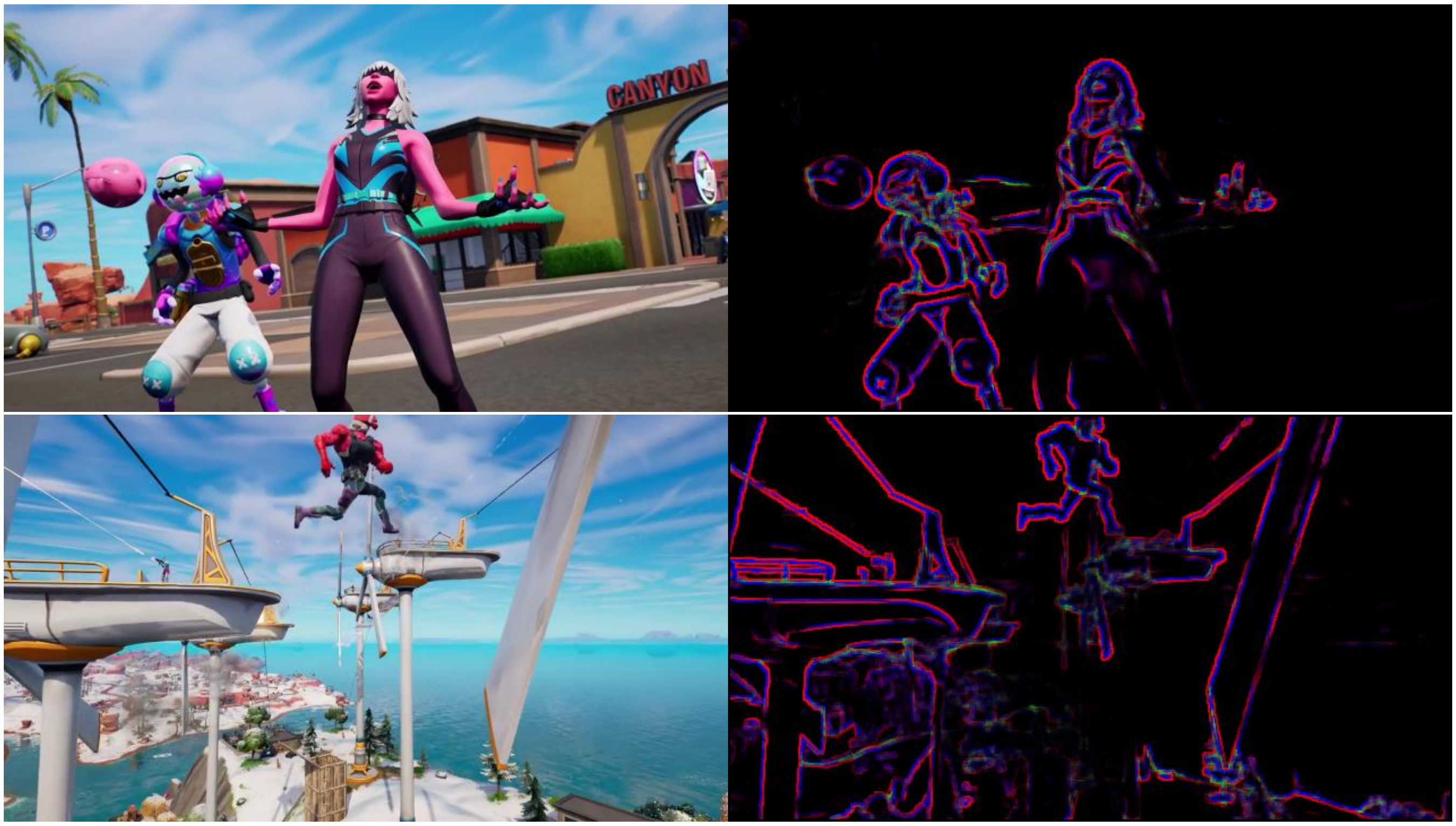

DAVIS:

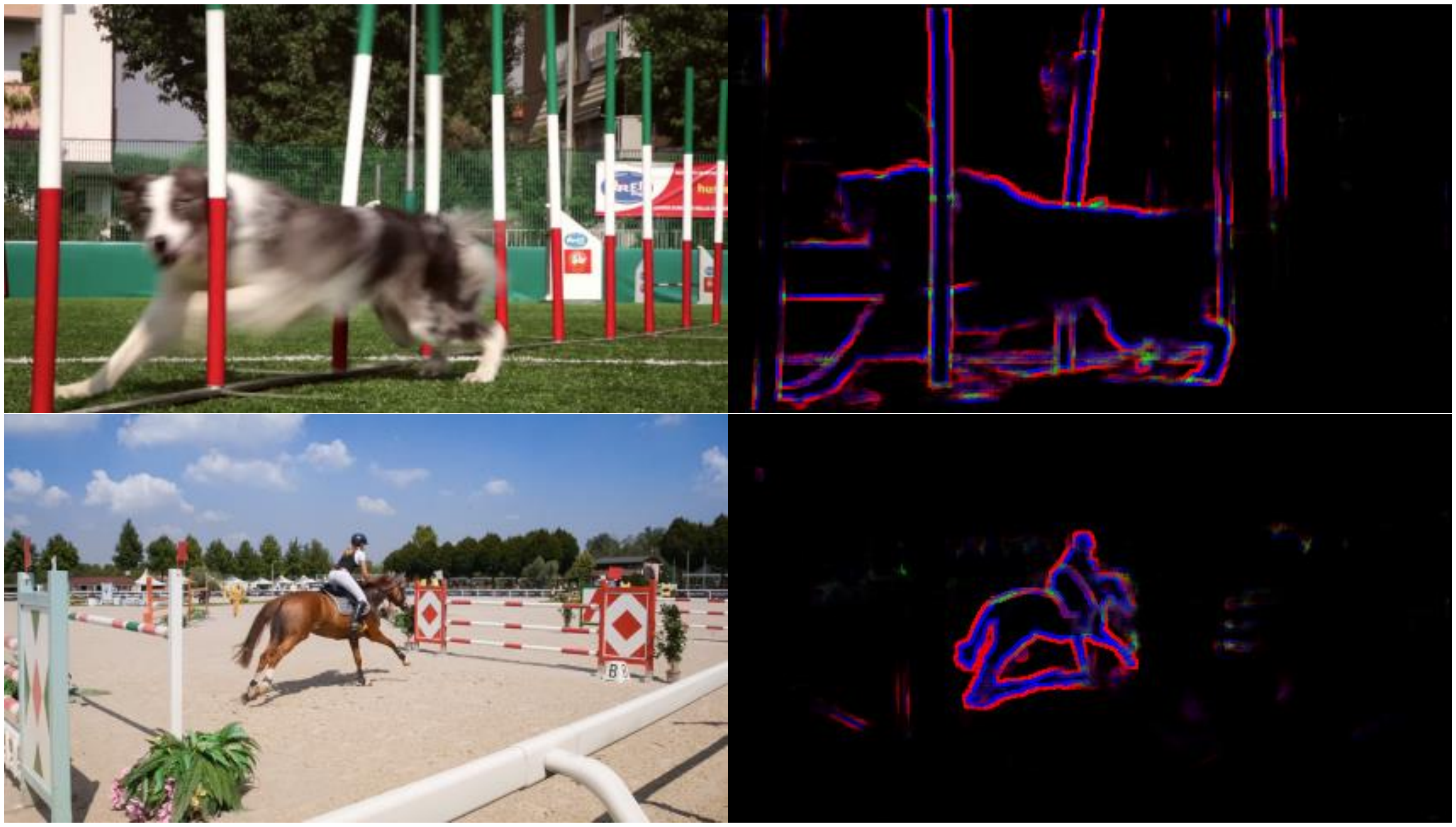

## D. MBSnet GradCAM examples

These visualizations provide a window into the breakdown of tasks as implemented by a trained MBSnet. The layout for each target follows the R(2+1)U-Net diagram of Figure 5, reproduced and annotated below (left) along with a key to the layout of the GradCAM maps (right):

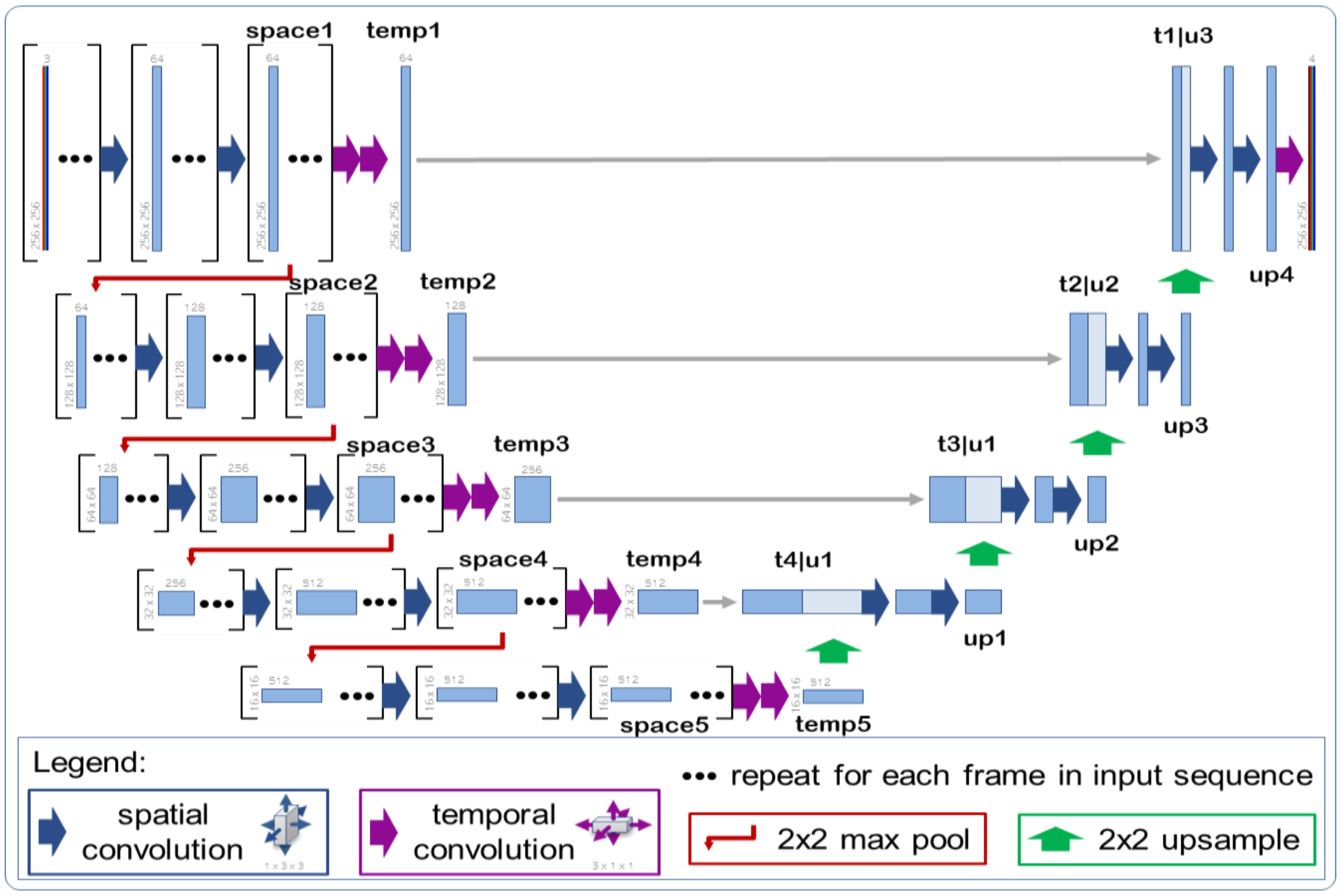

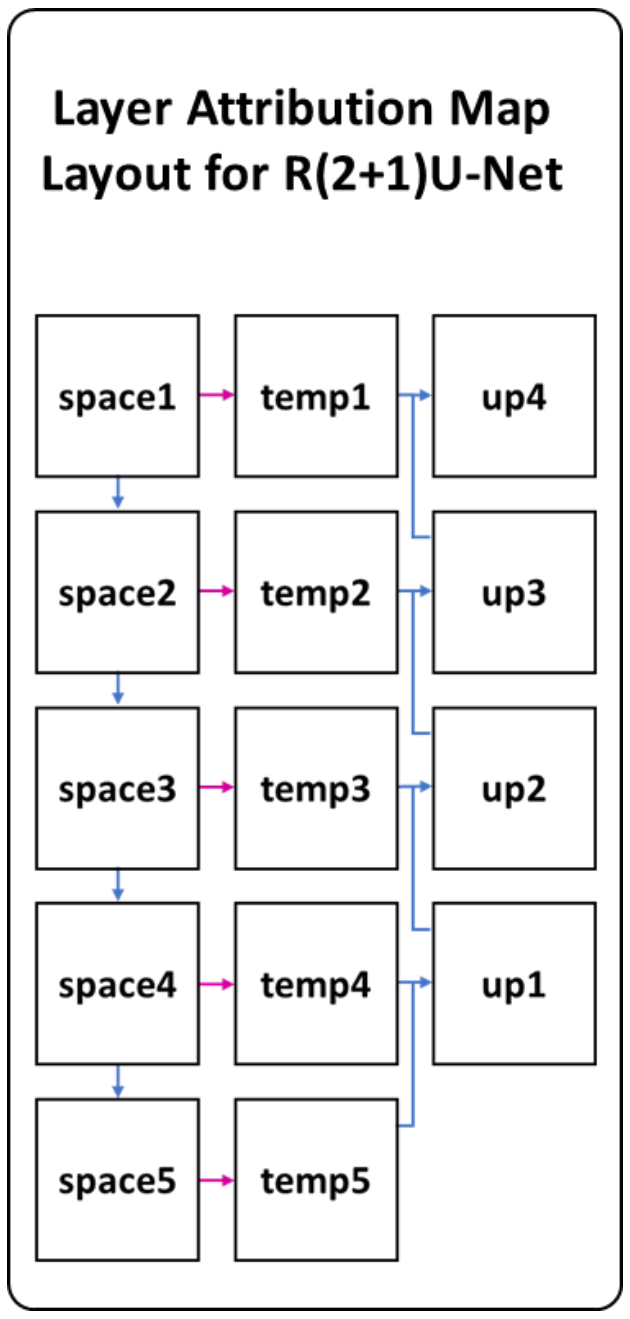

In this example, we illustrate the layer attribution maps for MBSnet inference of a single frame. Since the attribution changes for each class which gets predicted, we will show the maps for each class: 0 – Not part of a motion boundary; 1 – Foreground side of motion boundary; 2 – Background side of motion boundary; 3 – Both senses (i.e., foreground with respect to one boundary and background with respect to another). Note also that the spatial convolution outputs in the left column (space1, space2, …) actually represent a stack of individually spatially processed frames; we show only the center frame in the maps in this section. The temporal convolutions at each level produce a single temporal output frame corresponding to the center of the spatial sequence. The final MBSnet product, after one last convolution layer applied to up4, is the four band prediction map with softmax outputs for classes 0, 1, 2, and 3 at each pixel.

**input frame sequence (note variable blur, camera motion, and object velocity)**

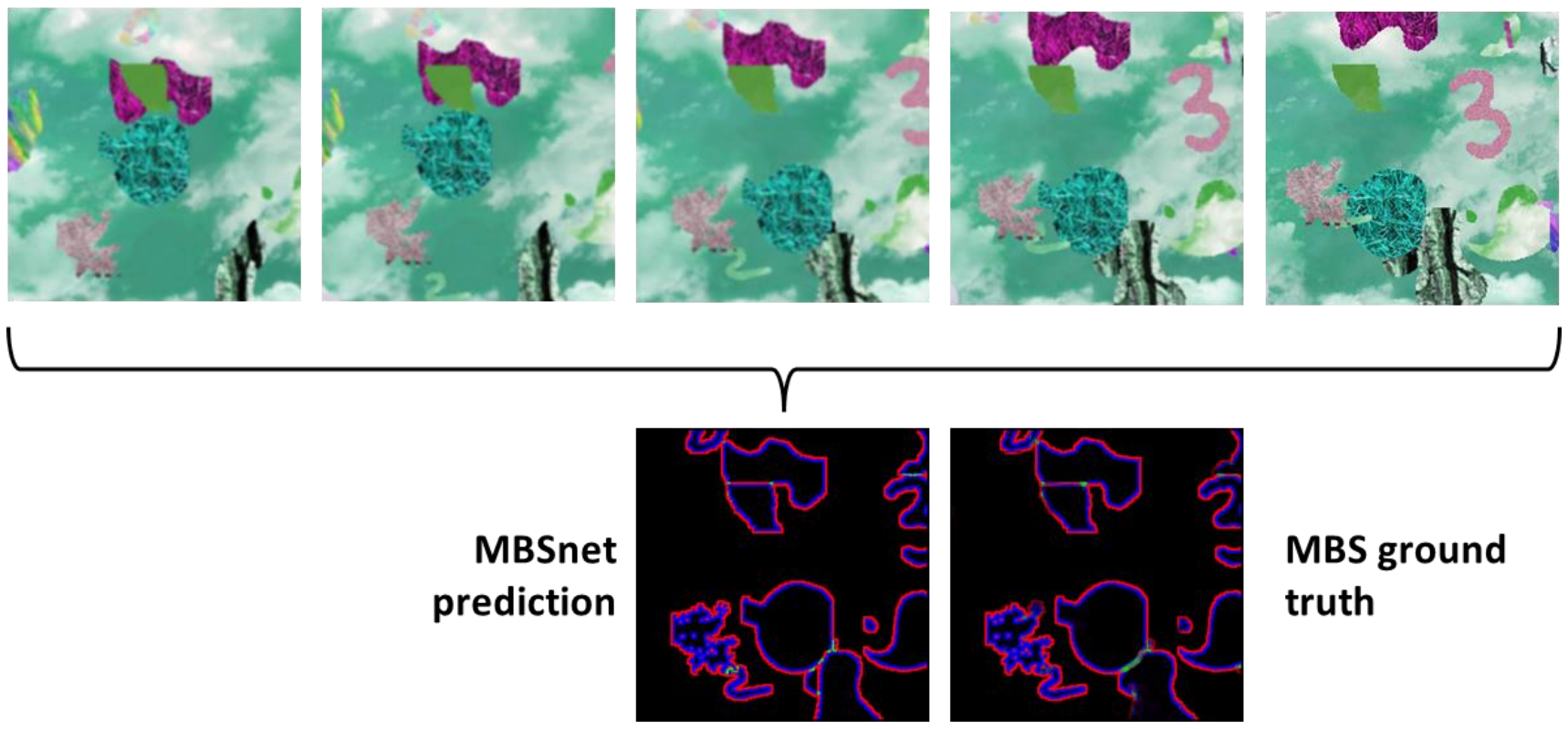

**MBSnet prediction**

**MBS ground truth**

In these MBSnet layer attribution maps that follow, we observe the following:

- Motion boundaries come into clear focus after the temporal convolutions
- The up2 layer for classes 1 and 2 demonstrate contrast in where the network focuses for foreground vs. background classes
- The final level seems to have marginal value for this task with this data.

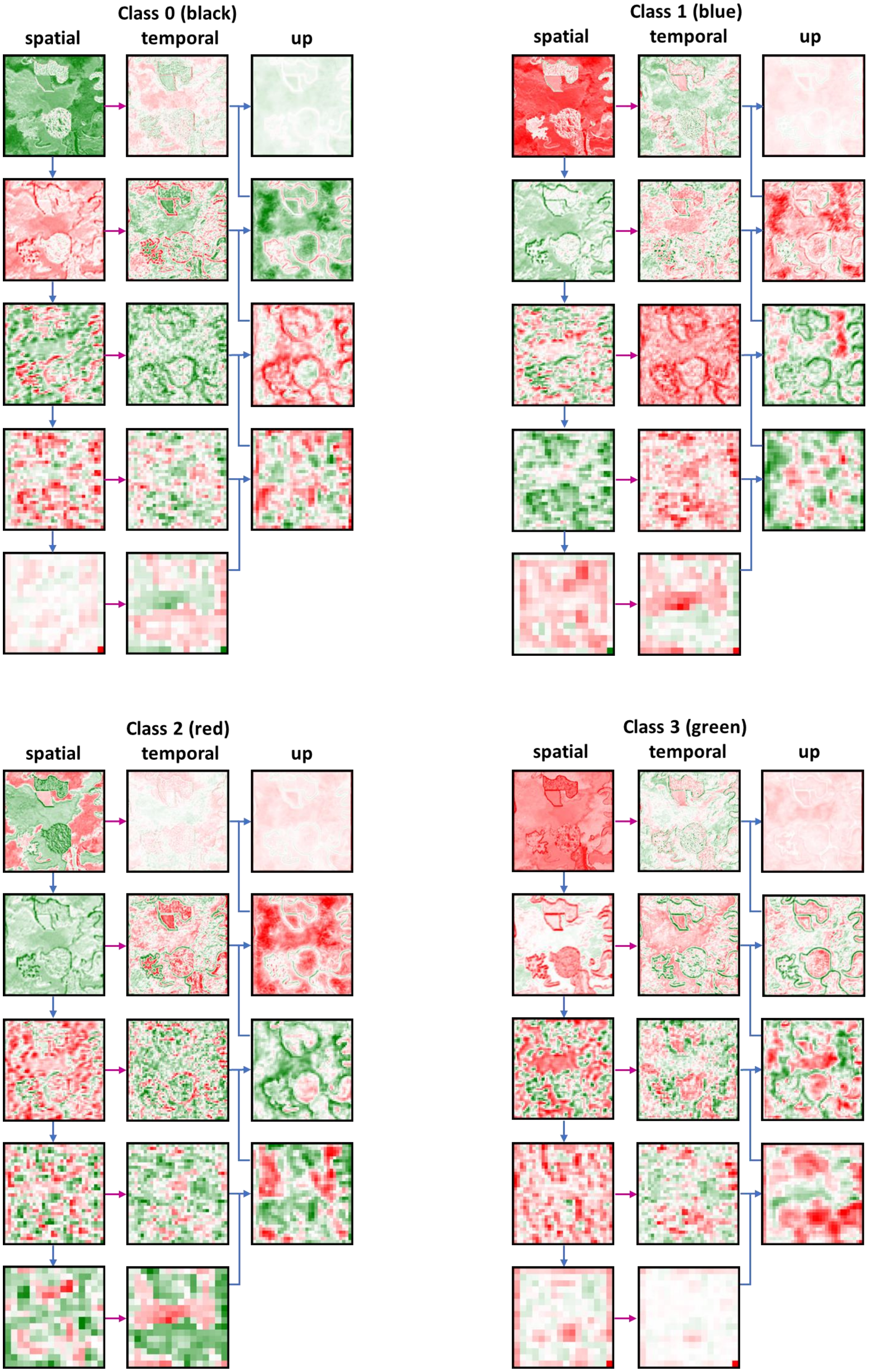

## E. ASPnet validation and test examples

Here we present a visual comparison of ASPnet performance on various generated scenes: (1) color splash texture only, (2) random noise texture only, (3) standard scene sampling, (4) standard scenes with temporal MBS dropout, and (5) standard scenes with temporal attention dropout.

When viewing these as snapshots, it is impossible to infer what is happening in the scene just by viewing the frame; we rely on the 'MBS truth' to shed light on which objects are moving and which are stationary.

1: Color Splash

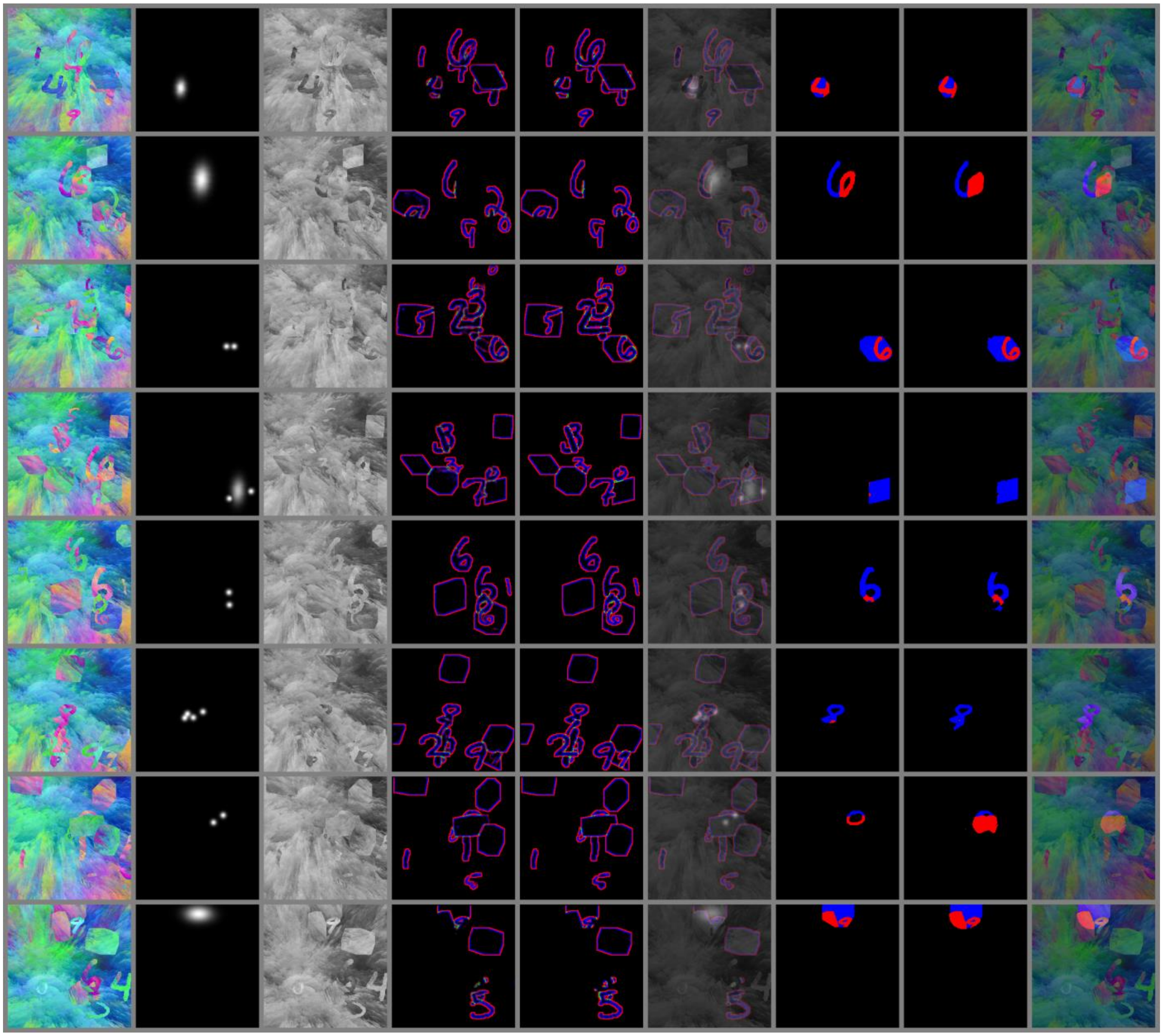

1) input frame | 2) attention signal | 3) input luma | 4) MBSnet prediction | 5) MBS ground truth | 6) ASP input (cols 2+3+4) | 7) ASP ground truth | 8) ASPnet prediction | 9) cols 1+8

2: Rand Noise

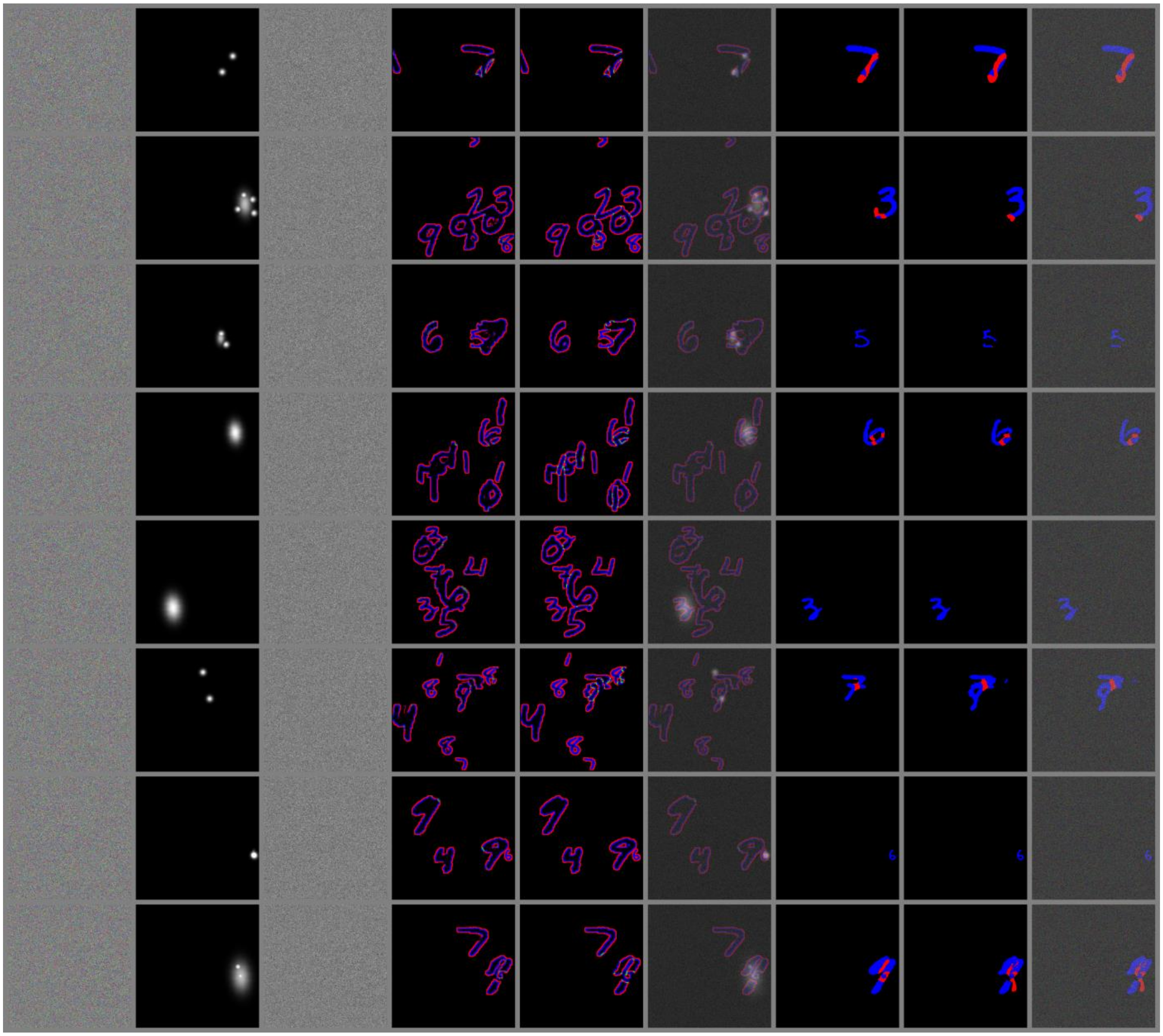

| 1) input frame | 2) attention signal | 3) input luma | 4) MBSnet prediction | 5) MBS ground truth | 6) ASP input (cols 2+3+4) | 7) ASP ground truth | 8) ASPnet prediction | 9) cols 1+8 |
|---|---|---|---|---|---|---|---|---|

3: Standard

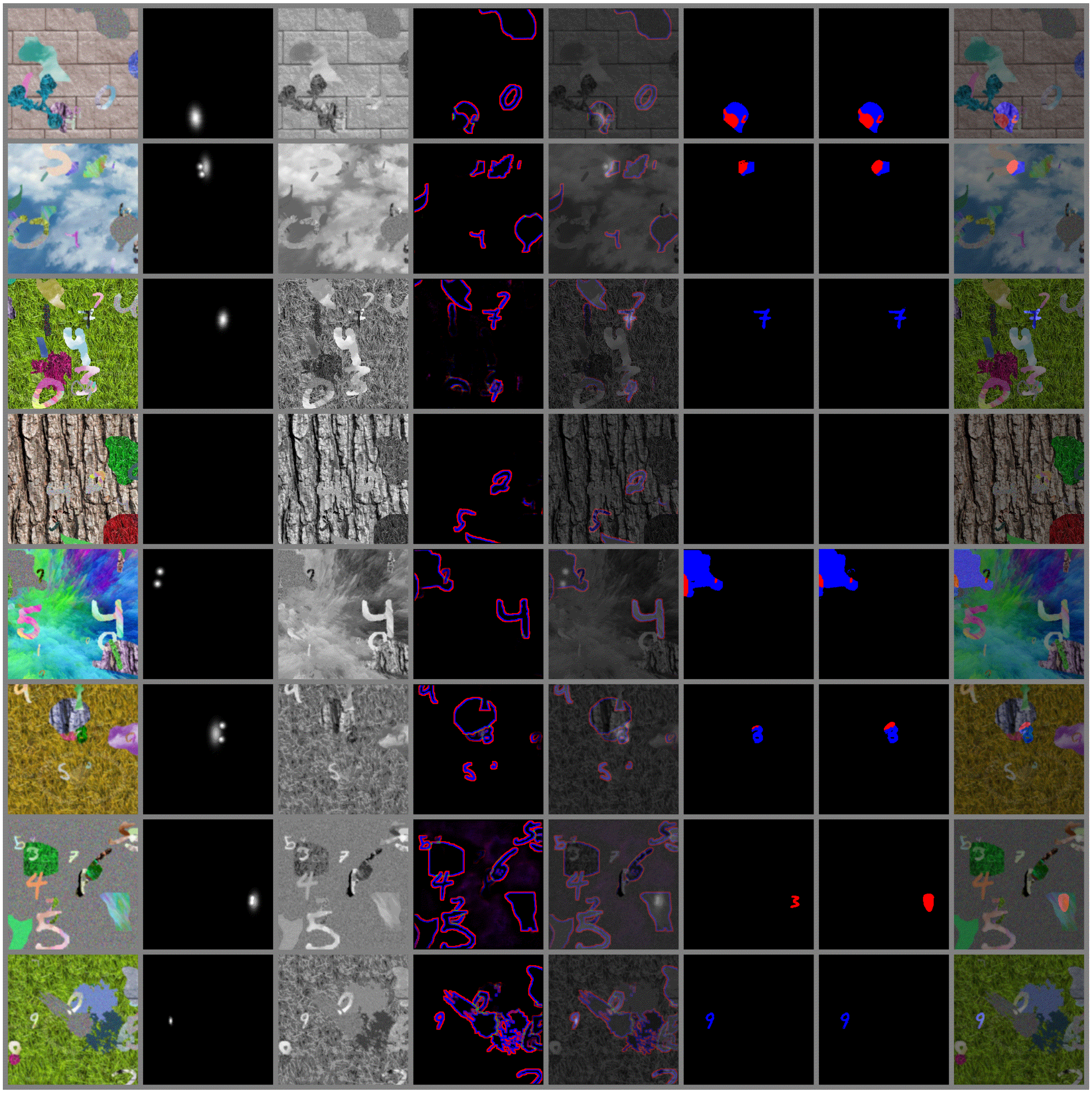

1) input frame 2) attention signal 3) input luma 4) MBSnet prediction 5) ASP input (cols 2+3+4) 6) ASP ground truth 7) ASPnet prediction 8) cols 1+7

4.a Temporal Dropout (Same set as previous)

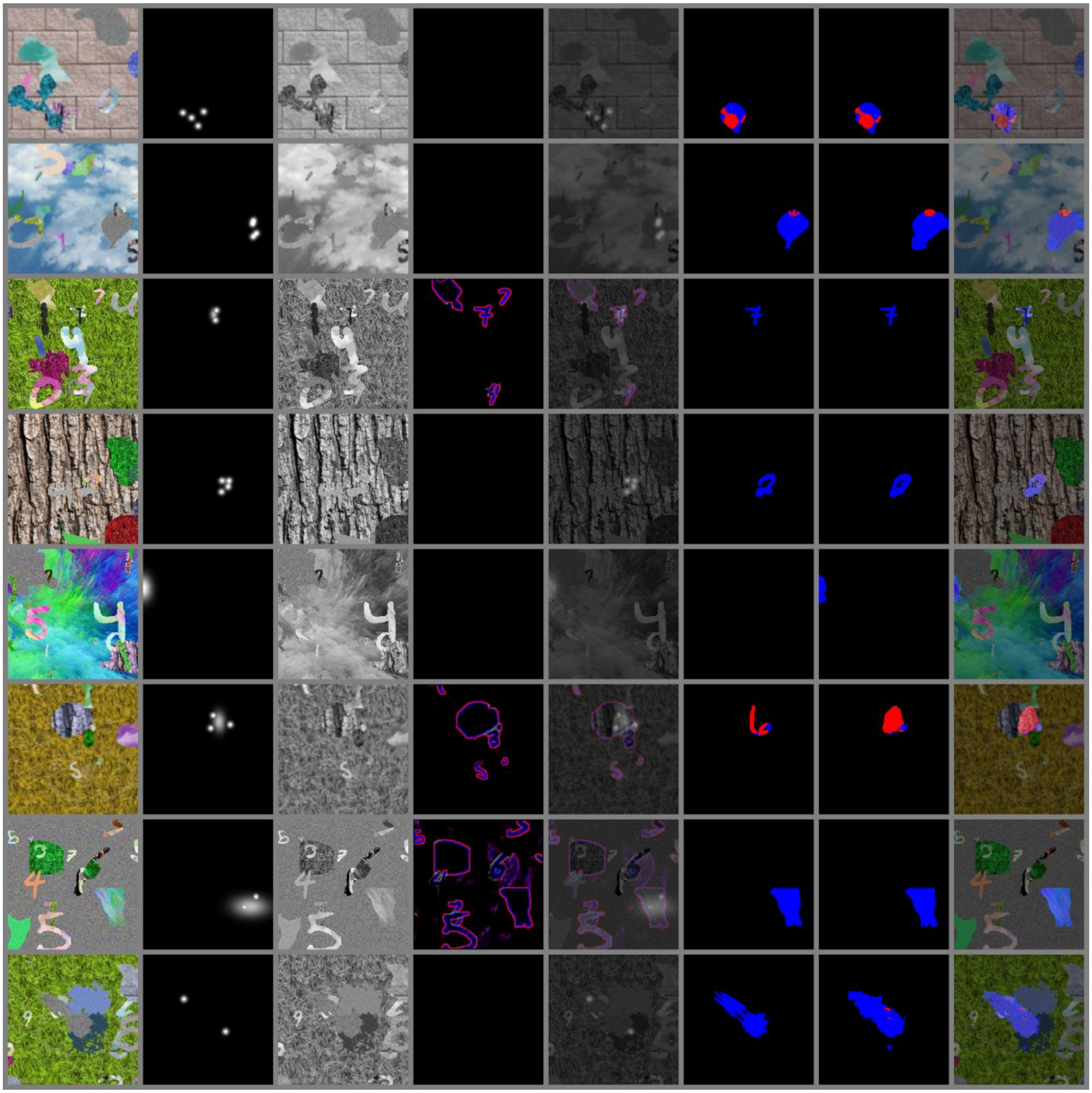

| 1) input frame | 2) attention signal | 3) input luma | 4) MBSnet prediction | 5) ASP input (cols 2+3+4) | 6) ASP ground truth | 7) ASPnet prediction | 8) cols 1+7 |
|---|---|---|---|---|---|---|---|

4.b Temporal Dropout (New set)

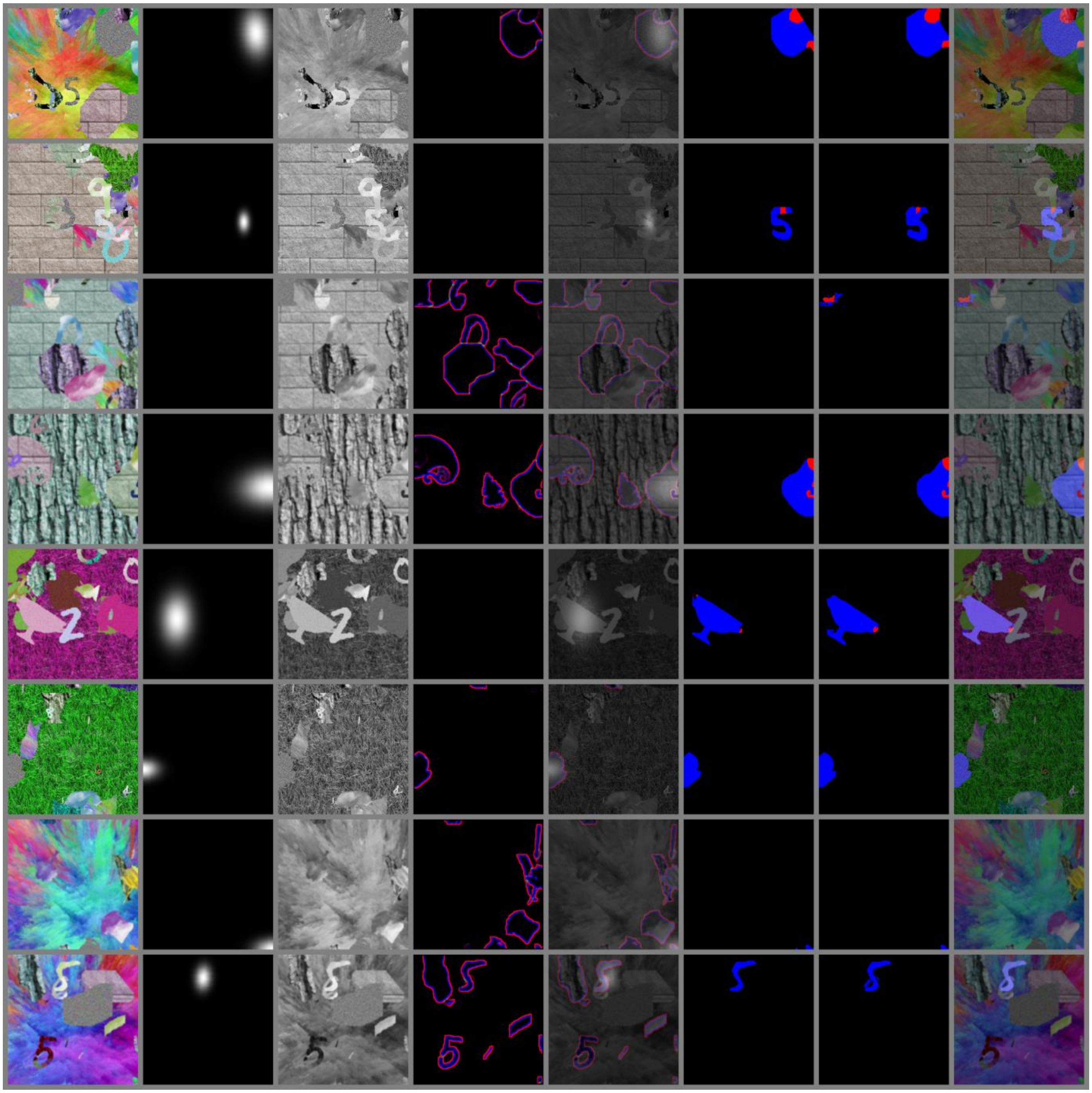

| 1) input frame | 2) attention signal | 3) input luma | 4) MBSnet prediction | 5) ASP input (cols 2+3+4) | 6) ASP ground truth | 7) ASPnet prediction | 8) cols 1+7 |
|---|---|---|---|---|---|---|---|

5: Attention Dropout

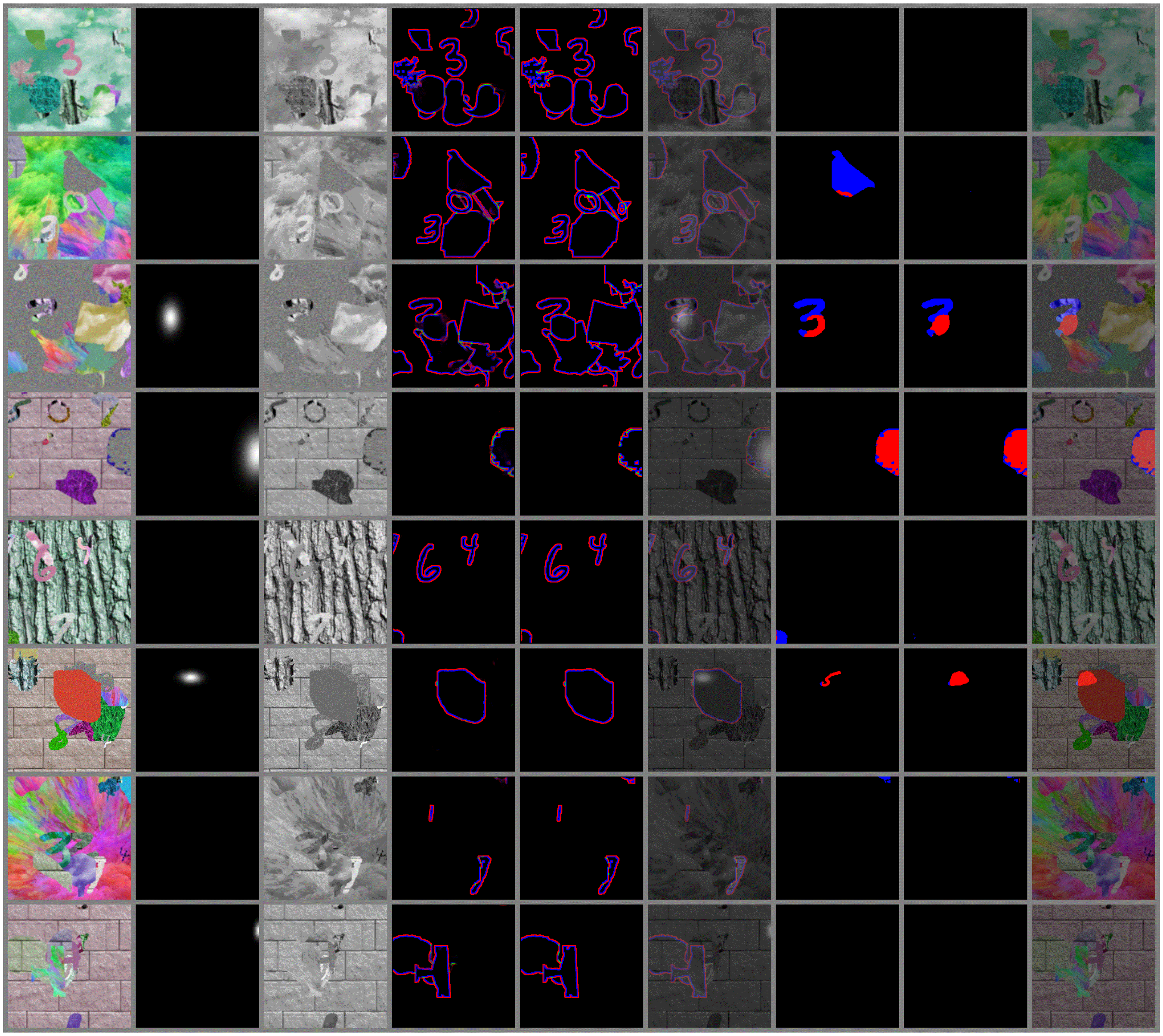

1) input frame 2) attention signal 3) input luma 4) MBSnet prediction 5) MBS ground truth 6) ASP input (cols 2+3+4) 7) ASP ground truth 8) ASPnet prediction 9) cols 1+8

## F. ASPnet GradCAM examples

These visualizations provide a window into the breakdown of tasks as implemented by a trained ASPnet. The layout for each target follows the R(2+1)U-Net diagram of Figure 5, reproduced and annotated below (left) along with a key to the layout of the GradCAM maps (right):

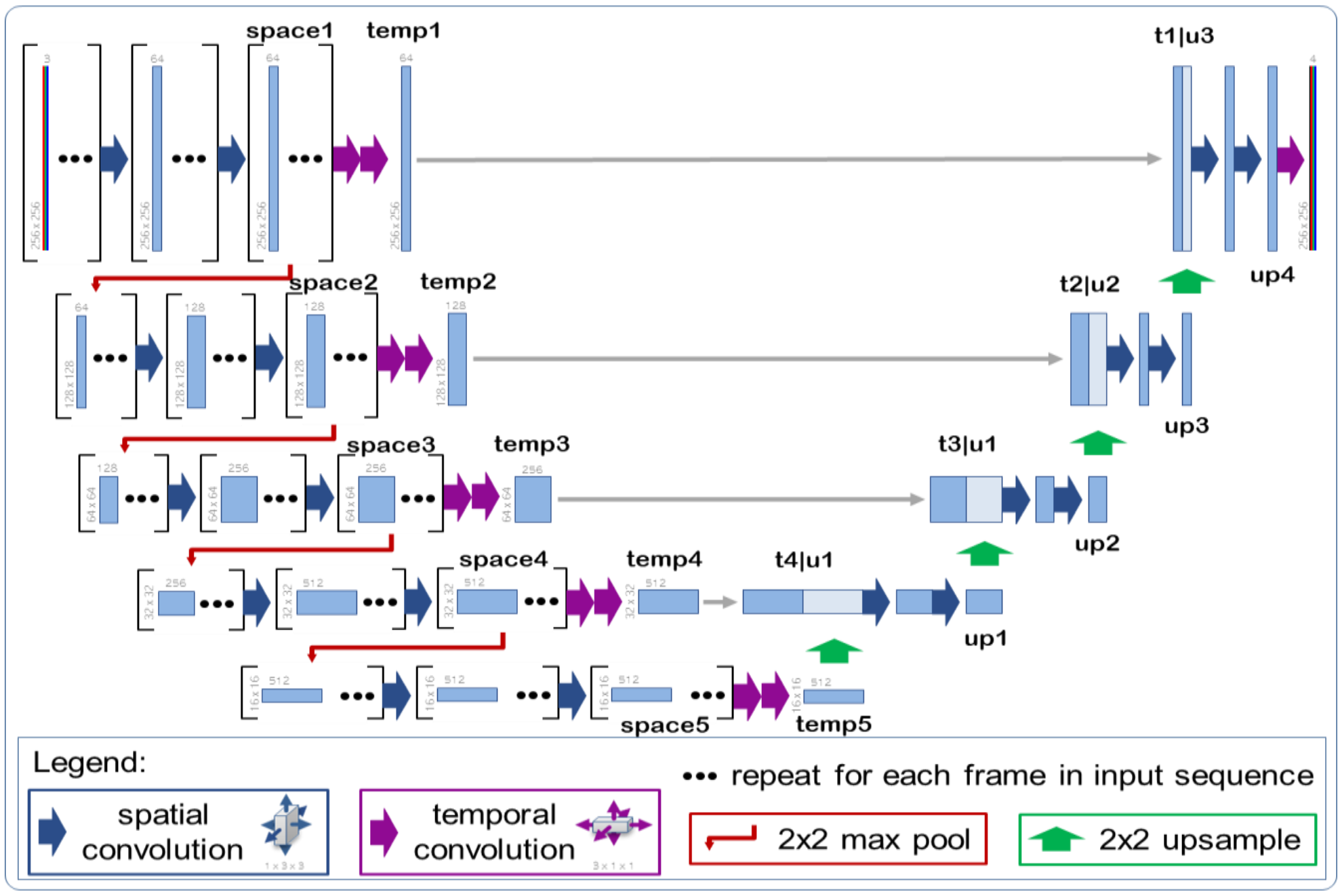

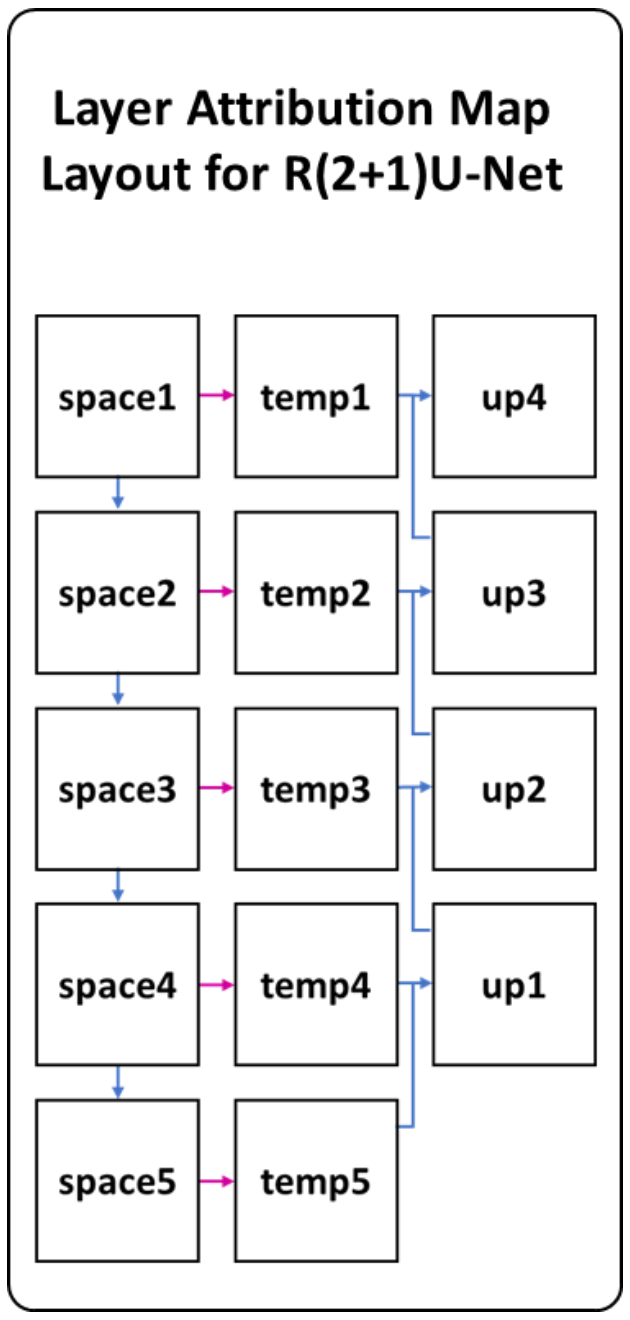

In the examples that follow, we illustrate the layer attribution maps for ASPnet inference of a single frame of a single attended object. Since the attribution changes for each class which gets predicted, we will show the maps for each class: 0 – Not the object; 1 – Visible object pixels; 2 – Obscured object pixels. Note also that the spatial convolution outputs in the left column (space1, space2, …) actually represent a stack of individually spatially processed frames. The temporal convolutions at each level (there are three for ASPnet, not the two that are shown here which represent MBSnet) produce a single temporal snapshot at the center of the current time window. The final ASPnet product, after one last convolution layer applied to up4, is the three band prediction map with softmax outputs for classes 0, 1, and 2 at each pixel.

Recall from Section 2.1 that the task of ASPnet is to “organize arrays of surfaces into unitary, bounded, and persisting objects”, one attended object at a time. GradCAM attributions give us some window into how ASPnet parses this task out amongst its various layers as shown above.

In these attribution maps, green denotes maximum impact on the eventual identification of pixels in the given class, while red is minimal and white is neutral. While a rigorous analysis of these outputs has not been performed, they do appear to provide insights into how the attention signal influences the output along multiple pathways, and also how the MBS information is exploited by the temporal convolutions to apparently suggest candidate surface bodies. While our future ASPnet implementation for more sophisticated objects may require more layers, it will still be a fairly compact architecture that is relatively amenable to interrogation and interpretation.

For this first example, we take an exhaustive look at the layer attributions for class 0 (“not this object”) pixels for the sequence pictured at right. We will follow five frames down the left side of the U-Net (the “space” layers from our layer attribution guide). Recall that an ASP input “frame” is the concatenation of the luma channel of the scene frame (though we show the full color frame here for interpretability), the three non-background bands of the MBSnet prediction frames, and the attention frame (i.e., heatmap).

The attended object is a rotated tree that has random dot texture

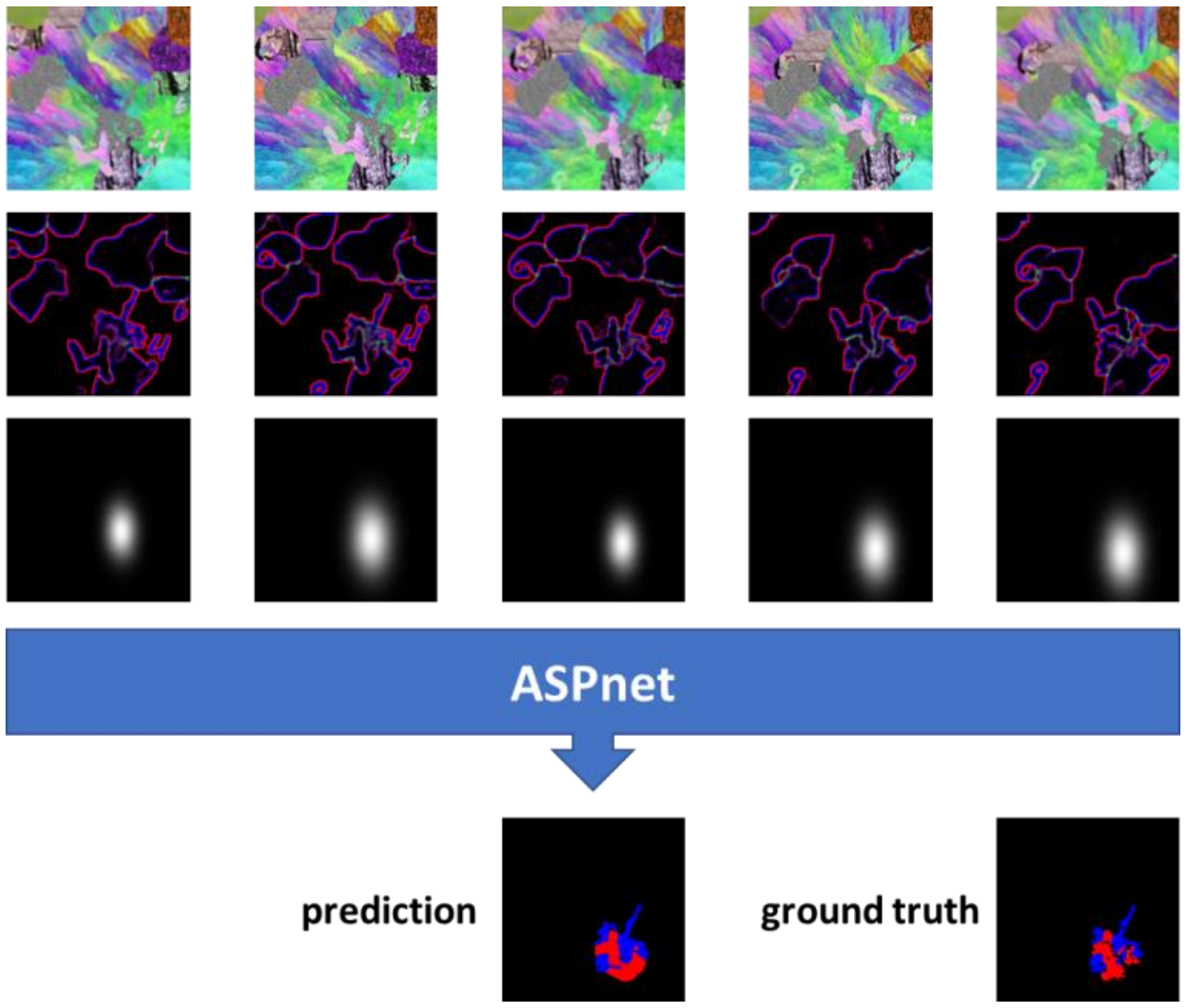

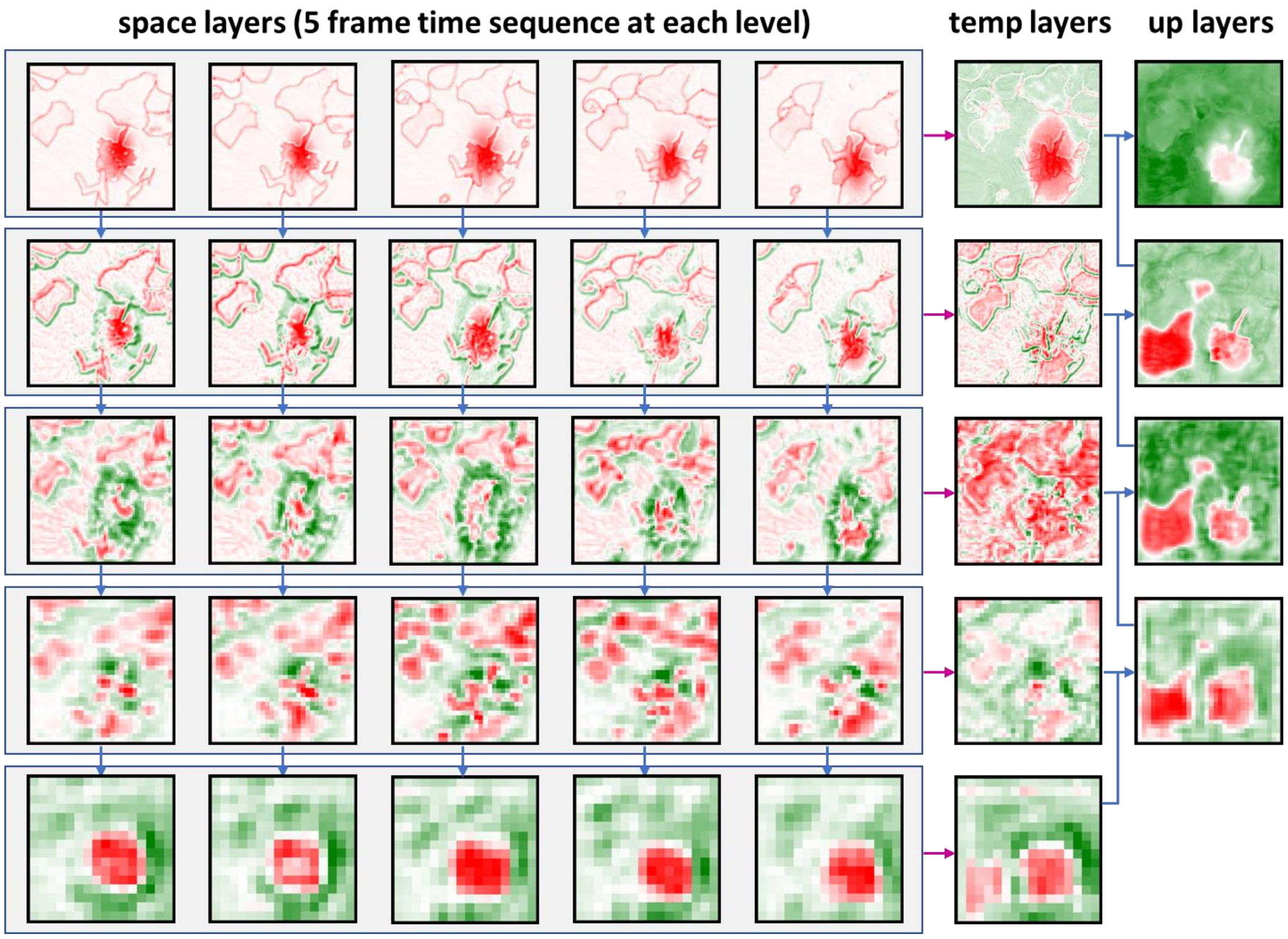

While much analysis remains to be done, we offer these initial observations (recall that this set of attribution maps is for ASP class 0):

- The top layer temporal convolution appears to aggregate and smear the attention sequence, and then leverage that as a final spatial filter on the product that emerges from the up layers
- In the space2 layer, objects have emerged from the MBS information, and those facing the attended region have high attribution
- In the space3 layer, the class 0 attributions (the deep green rings) are closing in around a more defined attended object candidate
- The culmination of the space layers, after temporal convolution, appears to be the complement of likely attended object candidates, a process which appears to leverage MBS information heavily, informed by attention

For class 1, the top layer takes a less generous interpretation of the spatial attention as the focus is on visible object pixels. We also see a somewhat different prioritization of various candidate surfaces (e.g., the green boundaries and then regions in space3 and space4, respectively):

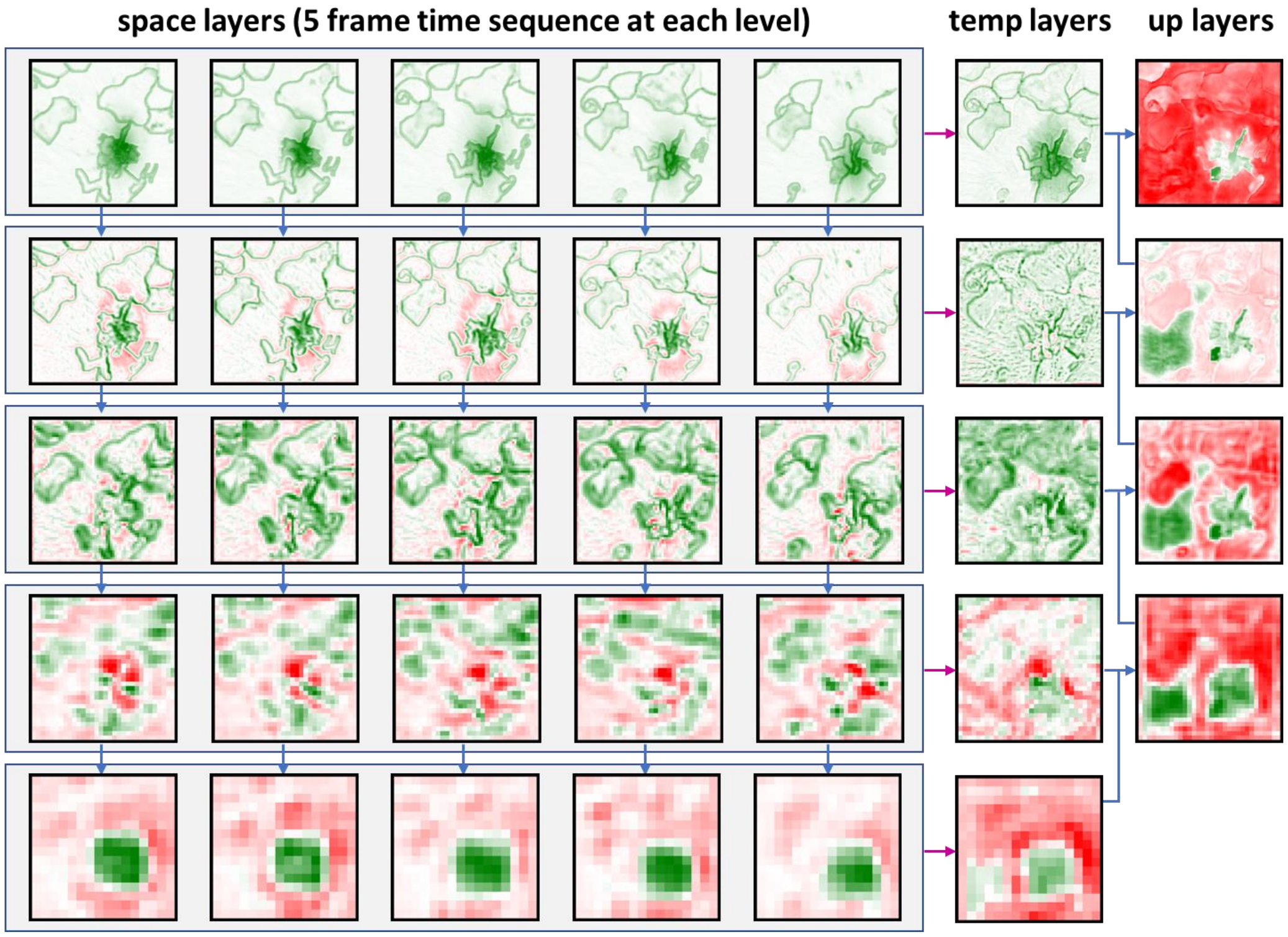

Finally class 2 appears to have a mix of the characteristics of the attributions from the other two classes:

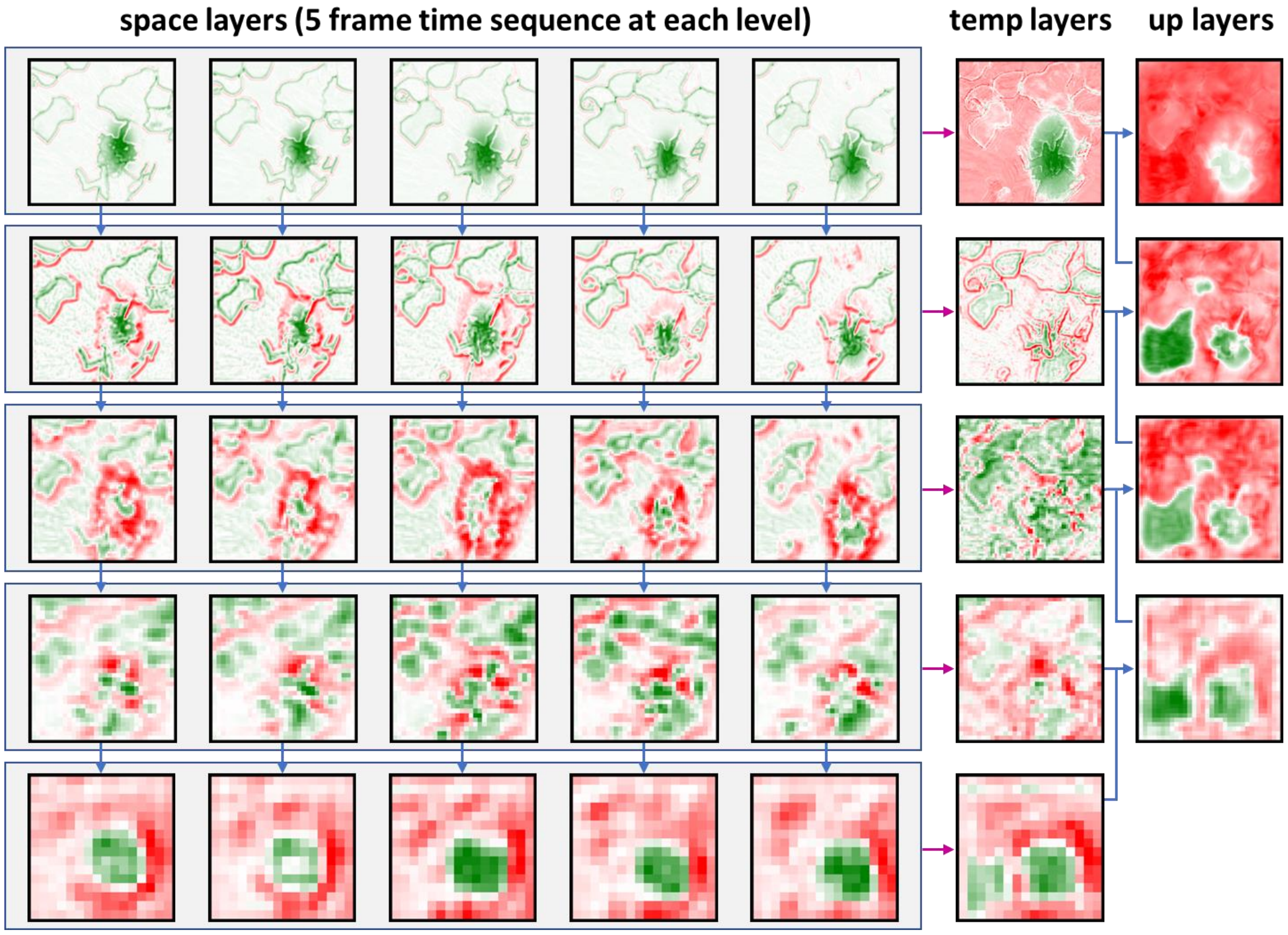

For this next example, we leave out the "space" layers, and instead we include three different time steps for the "temp" and "up" layers. We illustrate the input sequence just as for the previous example, but we include ASP predictions and ground truth corresponding to frames 1, 3 and 5 of the input sequence

The attended object is a vase that has "brick" texture that blends with the background. The attention sequence is in the form of a constellation of keypoints. Note that these keypoints at times overlap other objects (the stationary "2", the moving "7"), but that ASP has no problem grouping them with the surfaces that move together.

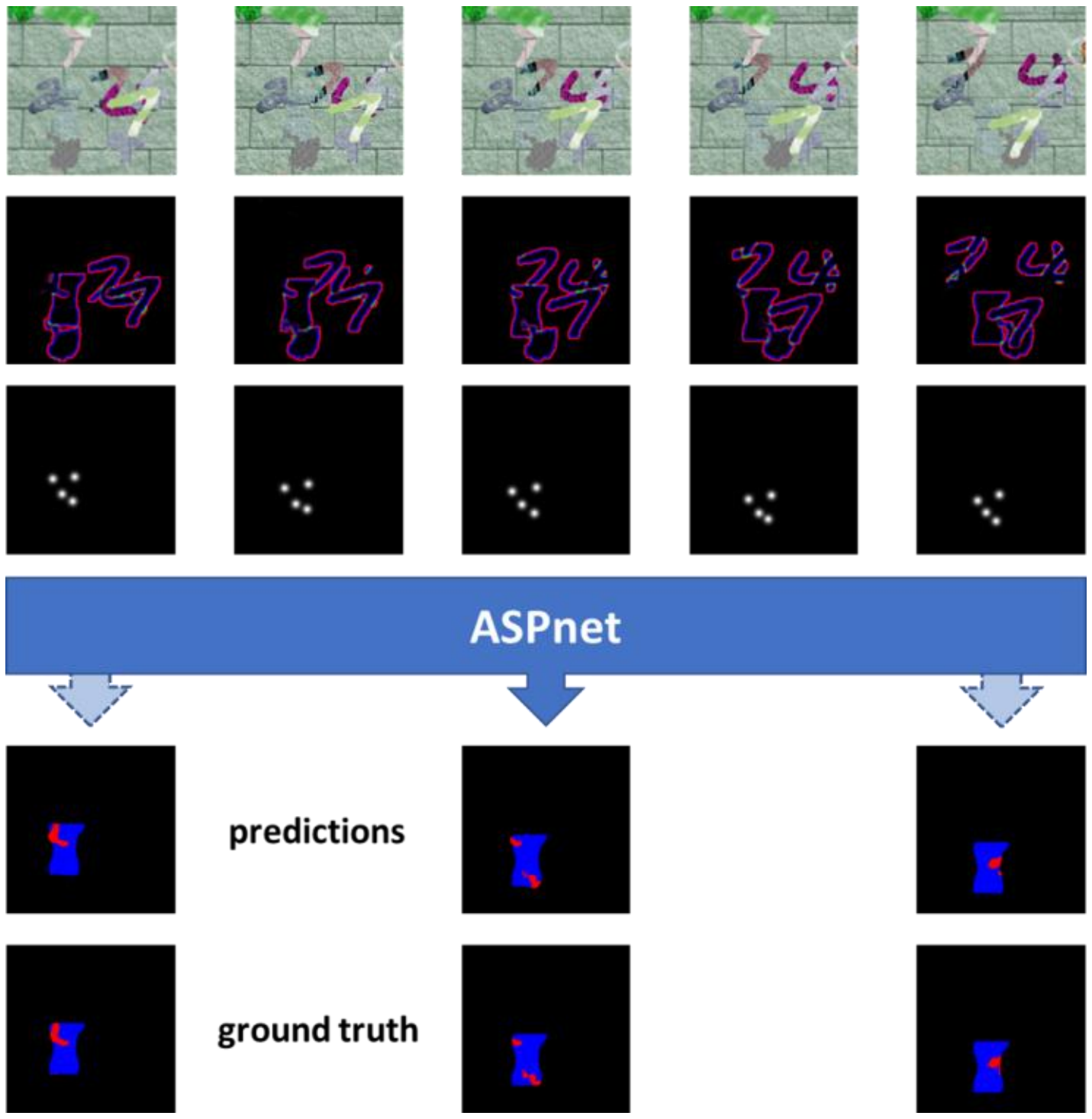

Here are the layer attribution maps for class 0. As before, the attention signal provides a strong inverse influence at the top layer, and proceeding down the temporal layers we see the building up of strong attribution around the attended object:

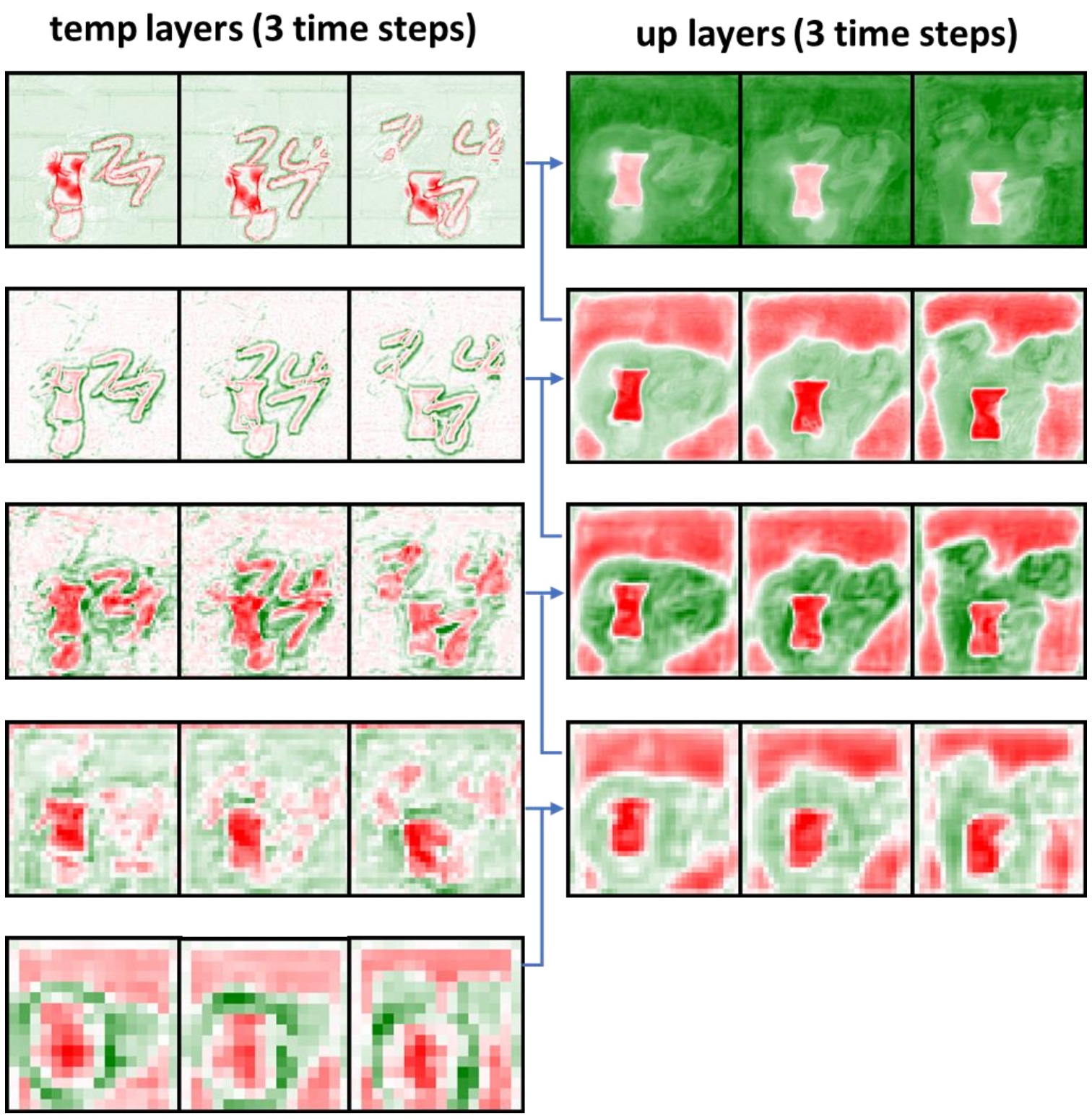

The class 1 “temp” layers are somewhat different, probably due to the lack of a central gaussian focus, but the “up” layers look similar:

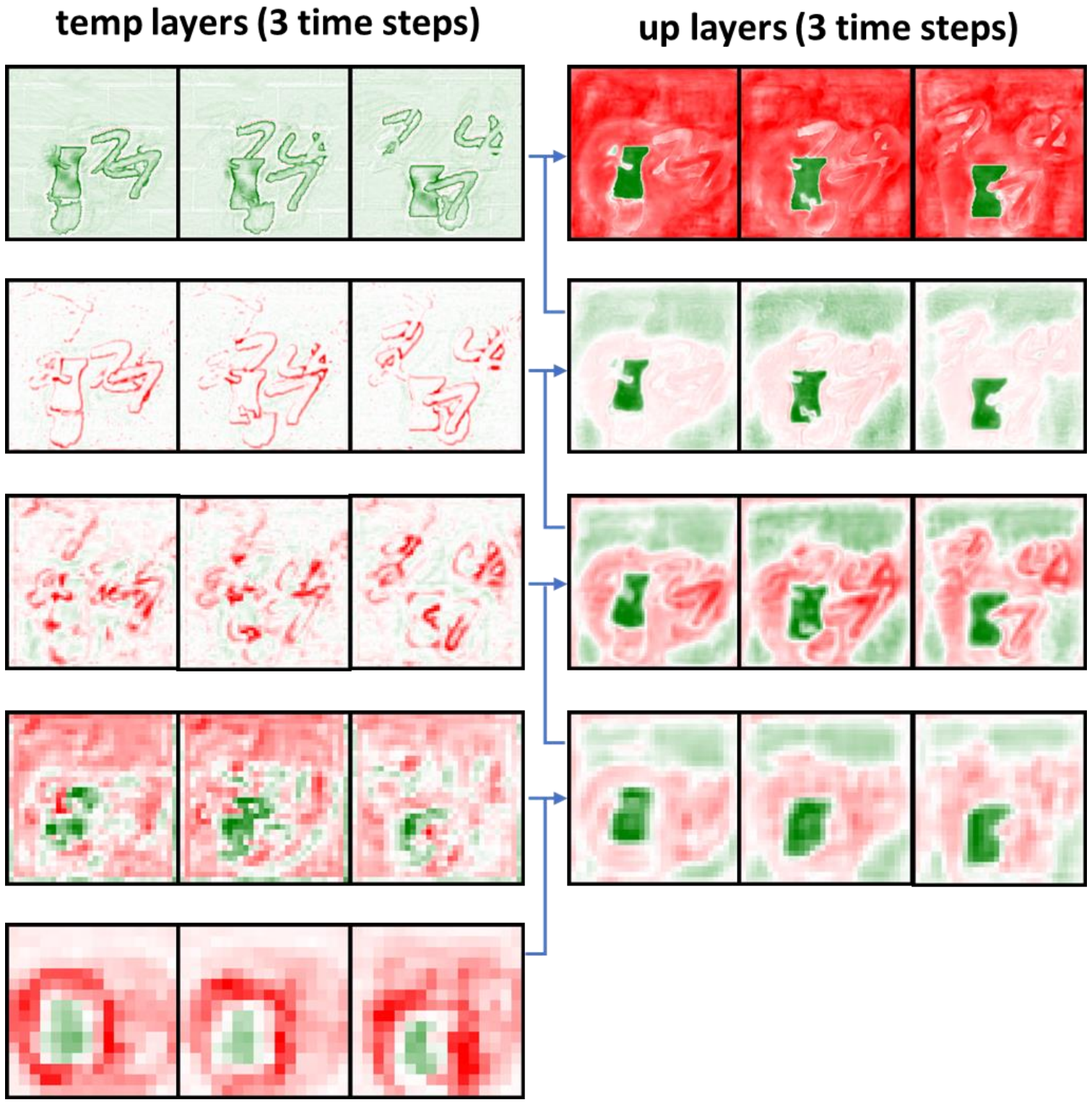

The second temporal layer of the class 2 maps are very interesting as we see strong attribution on all of the MBS green pixels (motion boundary intersections). This highlights the importance of these pixels as a strong indicator of possible obscuring surface intersections:

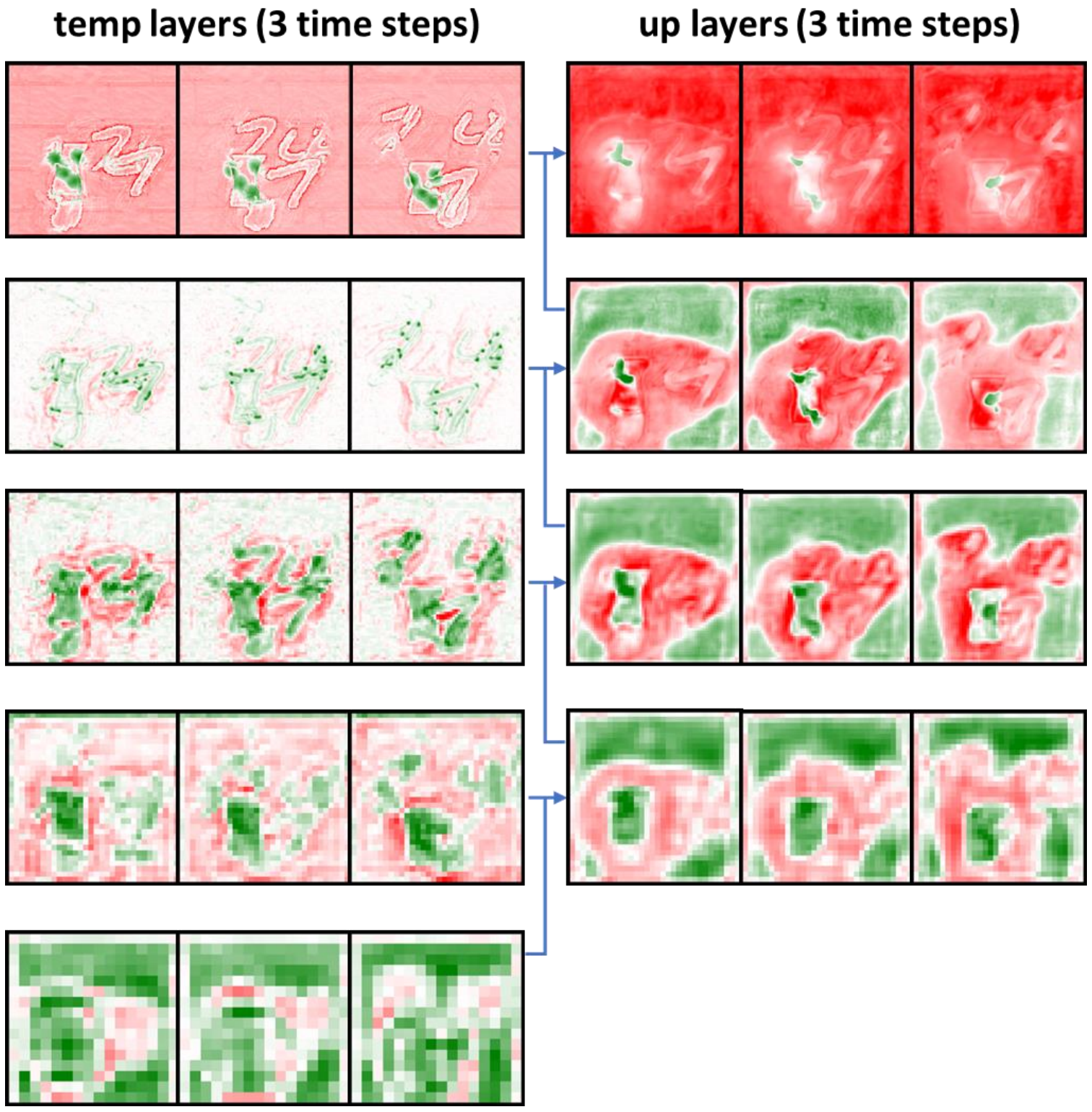